\documentclass[10pt,twocolumn,letterpaper]{article}

\usepackage{cvpr}

\usepackage{balance}

\usepackage{dblfloatfix}

\usepackage{grffile}

\usepackage{wrapfig}

\usepackage{placeins}

\usepackage[inline]{enumitem}

\usepackage{xargs}

\usepackage{xspace}

\usepackage{graphicx}
\usepackage{subcaption}

\usepackage{times}
\usepackage{epsfig}
\usepackage{amsmath,amssymb} % define this before the line numbering.

\usepackage{color}
\usepackage{bm}
\usepackage{mathtools}

\usepackage{tabu}

\usepackage{hyperref}
\usepackage[hyphenbreaks]{breakurl}

\usepackage[backend=biber,style=ieee,citestyle=numeric-comp]{biblatex}
\usepackage[utf8]{inputenc}

\usepackage{textcomp}

\usepackage{wrapfig}

\usepackage{adjustbox, multirow}

\newcommandx*{\skel}[1]{\ensuremath{#1}\xspace}
\newcommandx*{\ptThreeD}[2][1=i, 2=X, usedefault=@]{\ensuremath{\bm{#2}_{#1}}\xspace}
\newcommandx*{\makebold}[1]{\ensuremath{\bm{#1}}\xspace}
\newcommandx*{\plane}[1][1=\pi]{\makebold{#1}}
\newcommandx*{\normal}[1][1=n]{\ensuremath{\hat{\makebold{#1}}}\xspace}
\newcommandx*{\bmu}{\makebold{\mu}}
\newcommandx*{\symThreeD}{\skel{s_\text{3D}}}
\newcommandx*{\tThreeD}{\skel{t_\text{3D}}}
\newcommandx*{\symTwoD}{\skel{s_\text{2D}}}
\newcommandx*{\tTwoD}{\skel{t_\text{2D}}}

\newlength{\savelength}
\newenvironmentx*{tighttabular}[1]%
{%
  \setlength{\savelength}{\tabcolsep}%
  \setlength{\tabcolsep}{0pt}
  \begin{tabular}{#1}%
    }%
    {\end{tabular}%
    \setlength{\tabcolsep}{\savelength}%
  }

\newcommand{\figref}[1]{Figure~\ref{#1}}
\newcommand{\tabref}[1]{Table~\ref{#1}}
\newcommand{\secref}[1]{Section~\ref{#1}}

\hypersetup{
  pagebackref=true,
  breaklinks=true,
  colorlinks=true,
}

\begin{document}

% \cvprfinalcopy % *** Uncomment this line for the final submission

% \def\iccvPaperID{4230} % *** Enter the ICCV Paper ID here
% \def\httilde{\mbox{\tt\raisebox{-.5ex}{\symbol{126}}}}

% % Pages are numbered in submission mode, and unnumbered in camera-ready
% \ificcvfinal\pagestyle{empty}\fi

\cvprfinalcopy % *** Uncomment this line for the final submission

\def\cvprPaperID{2598} % *** Enter the CVPR Paper ID here
\def\httilde{\mbox{\tt\raisebox{-.5ex}{\symbol{126}}}}

% Pages are numbered in submission mode, and unnumbered in camera-ready
% \setcounter{page}{1}

%%%%%%%%% TITLE
% \title{3D Face Synthesis for Facial Analysis Tasks, How Much Impact?}
% \title{Improving Robustness of Face Recognition Systems with Data Synthesis using 3D Face Modeling}
% \title{Comprehensive Data Synthesis for Improving Performance of Face Recognition Systems through 3D Face Modeling}
% \title{Bells and Whistles for 3D-Aided Face Analysis}
\title{3D-Aided Data Augmentation for Robust Face Understanding}

\author{Yifan Xing \hspace{2cm} Yuanjun Xiong \hspace{2cm} Wei Xia \\
  AWS/Amazon AI \\
  {\tt\small yifax, yuanjx, wxia@amazon.com}
}

% \and

\maketitle

\ifcvprfinal\pagestyle{empty}\fi
% \thispagestyle{fancy}

%%% Local Variables:
%%% mode: latex
%%% TeX-master: "data_augmentation_for_face_recognition"
%%% End:

% \begin{abstract}
%   %%%%%%%%%%%%%%%%%%%%%%%%%%%%%%%%%%%%%%%%%%%%%%%%%%%%%%%
%   % Structured abstract must have:
%   % Background
%   % Methods
%   % Results
%   % Conclusion
%   %%%%%%%%%%%%%%%%%%%%%%%%%%%%%%%%%%%%%%%%%%%%%%%%%%%%%%%
%   % Background
%   We have implemented a system for augmenting face-recognition and face-verification datasets.
%   % Methods
%   The system is composed of \begin{enumerate*}[label=(\roman*)]
%     \item a 3D-reconstruction module, which estimates the geometry of a subject's face from a single input image;
%     \item a pose-estimation module, which estimates the orientation of the face with respect to the viewing camera, assuming orthographic projection;
%     \item a module for the ray tracing of 2D face landmarks onto the reconstructed 3D surface;
%     \item and a rendering engine to produce novel viewpoints of the face depicted in the input image.
%     \end{enumerate*}
%   % Results
%     The proposed system produces realistic renderings of the input image from multiple viewpoints, each associated with geometrically accurate of locations of landmarks annotated in the original view.
%   % Conclusion
%     %{\color{red} CONCLUSION.}
% \end{abstract}

\begin{abstract}
Data augmentation has been highly effective in narrowing the data gap and reducing the cost for human annotation, especially for tasks where ground truth labels are difficult and expensive to acquire. In face recognition, large pose and illumination variation of face images has been a key factor for performance degradation. However, human annotation for the various face understanding tasks including face landmark localization, face attributes classification and face recognition under these challenging scenarios are highly costly to acquire. Therefore, it would be desirable to perform data augmentation for these cases. But simple 2D data augmentation techniques on the image domain are not able to satisfy the requirement of these challenging cases. As such, 3D face modeling, in particular, single image 3D face modeling, stands a feasible solution for these challenging conditions beyond 2D based data augmentation. To this end, we propose a method that produces realistic 3D augmented images from multiple viewpoints with different illumination conditions through 3D face modeling, each associated with geometrically accurate face landmarks, attributes and identity information. Experiments demonstrate that the proposed 3D data augmentation method significantly improves the performance and robustness of various face understanding tasks while achieving state-of-arts on multiple benchmarks.
\end{abstract}

%%% Local Variables:
%%% mode: latex
%%% TeX-master: "data_augmentation_for_face_recognition"
%%% End:

\section{Introduction\label{sec:introduction}}

Image based human face understanding systems aim to recognize from the input face image the face's identity and persistent attributes,~\emph{e.g.} age and gender. When applied in real-world scenarios, these systems are expected to be robust against variations in pose and illumination.
In the past decade, the emergence of deep learning based models, trained with large-scale face datasets~\cite{guo2016msceleb, Cao18, wang2018devil, sagonas2013300, bulat2017far, liu2015deep}, has greatly boosted the accuracy of face understanding systems~\cite{wen2016discriminative, DBLP:journals/corr/abs-1809-07586, Cao18, wang2018devil} especially for near frontal faces under normal lighting conditions.
However, the robustness of the face understanding models to large variations is still unsatisfying.
Face recognition accuracy and face attributes detection accuracy fall significantly when profile faces and uncommon lighting conditions are present~\cite{Cao_2018_CVPR_pose_equivariant, Zhao_2018_CVPR, Ha2018wacv, zhao20183d, Guo2019PFLDAP, yin2018multi}. %cite the papers showing face recognition systems have low accuracies for profile faces.

\begin{figure}[t]
%   \begin{subfigure}[tbh]{0.49\linewidth}
%     \includegraphics[width=1.\columnwidth]{./Figures/ms1m_imgs_with_aug_prob_0.4_hist_yaw}
%     \caption{}\label{fig:pos_plot_gender}
%   \end{subfigure}
%     \begin{subfigure}[tbh]{0.49\linewidth}
%     \includegraphics[width=1.\columnwidth]{./Figures/celeba_before_after_hist_yaw}
%     \caption{}\label{fig:pos_plot_celeba}
%   \end{subfigure}
  \begin{subfigure}[tbh]{1.\linewidth}
    \includegraphics[width=1.\columnwidth]{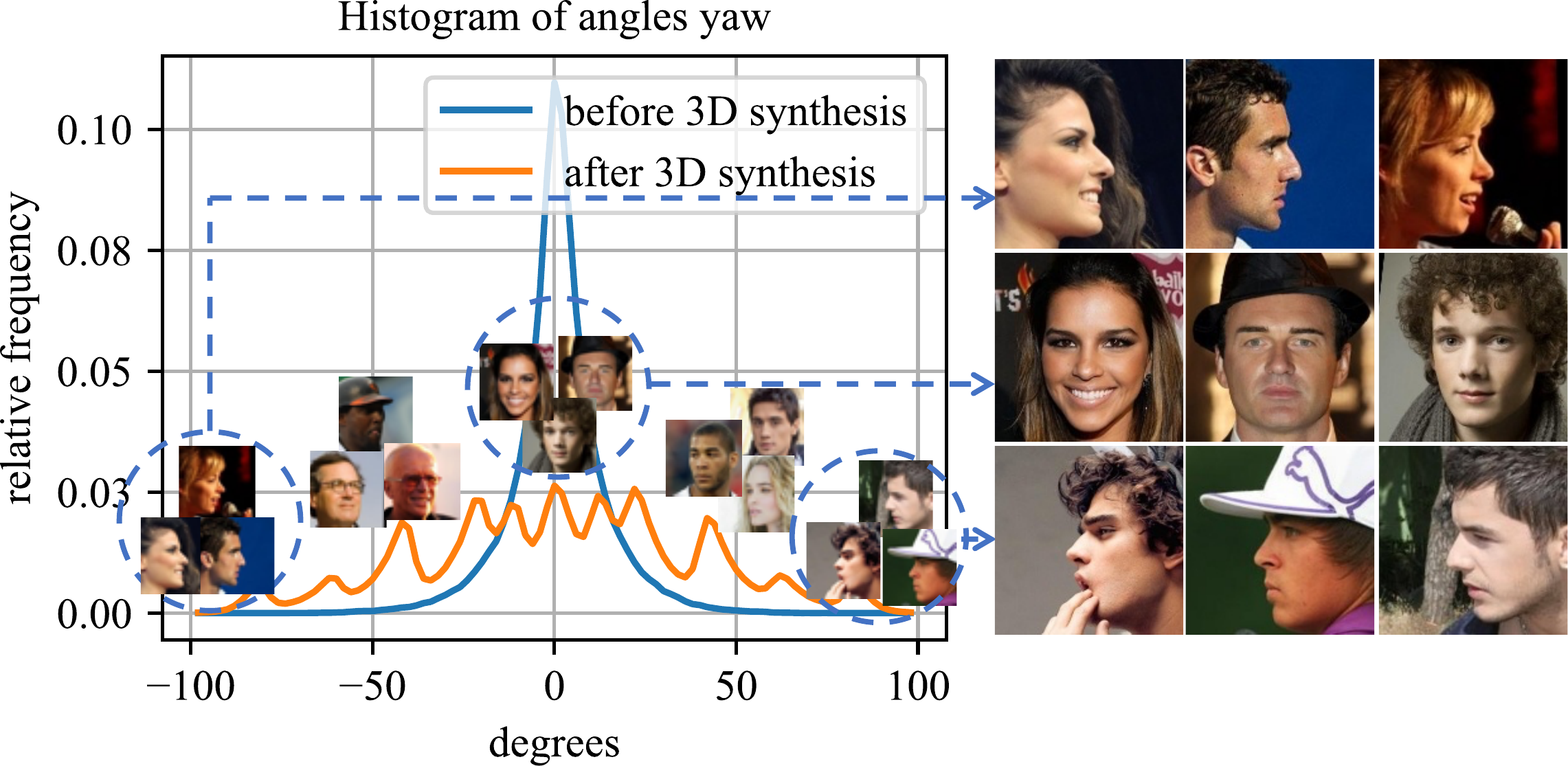}
  \end{subfigure}
  
  \caption{\label{fig:teaser_plots} Histogram of yaw angles of face images before (blue) and after (yellow) applying the proposed 3D face data synthesis method on the CelebA \cite{liu2015deep} dataset; Face images overlaid on the histogram illustrate the yaw pose distribution; Images on the right: 1st, 2nd and 3rd row represent -90, 0 and 90 degree yaw angles respectively.}
\end{figure}

\begin{figure}[t]
\setlength{\tabcolsep}{1pt}
    \begin{subfigure}[tbh]{0.245\linewidth}
    \includegraphics[width=1.\columnwidth]{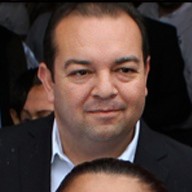}
    \caption{original}\label{fig:illumination_plot}
  \end{subfigure}\hfill%
    \begin{subfigure}[tbh]{0.245\linewidth}
    \includegraphics[width=1.\columnwidth]{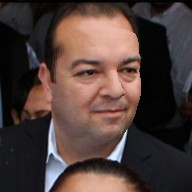}
    \caption{yaw rotated}\label{fig:illumination_plot}
  \end{subfigure}\hfill%
    \begin{subfigure}[tbh]{0.245\linewidth}
    \includegraphics[width=1.\columnwidth]{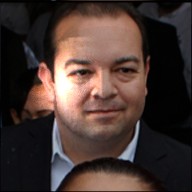}
    \caption{relighted}\label{fig:illumination_plot}
  \end{subfigure}\hfill%
    \begin{subfigure}[tbh]{0.245\linewidth}
    \includegraphics[width=1.\columnwidth]{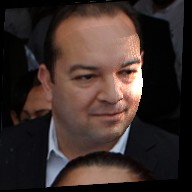}
    \caption{rotate relight}\label{fig:illumination_plot}
  \end{subfigure}
%   \vspace{-1em}
  \caption{\label{fig:teaser_fig_aug} Exemplar 3D augmented images: (a) original, (b) yaw rotated, (c) relighted and (d) rotated + relighted.}
\end{figure}

% \begin{figure}[t]
% \setlength{\tabcolsep}{1pt}

%   \vspace{-1em}
%   \caption{\label{fig:teaser_fig_aug} }
% \end{figure}

We argue it is due to two major reasons. 
First, although the existing datasets with millions of face images are great resources for learning models that can extract effective face representation, the variations of pose and lighting is lacking. 
Thus the models learned from these datasets have difficulties extracting face features from profile or ill-illuminated images.
\figref{fig:teaser_plots} illustrates the pose distribution of one popular face dataset.
Second, due to the difficulty of manually annotating groudtruth data on faces with large poses and lighting, the training data also has increasing labeling noise for this subset of images.
To achieve robust face understanding against such variation, it requires a method that can generate a ~\emph{large volume} of~\emph{high quality} training data with accurate annotations.

Recent advances in single image 3D face modeling approaches provide a unique opportunity to achieve the above goal. In this work, we propose a unified pipeline built on 3D face modeling to generate training data from existing face images. The proposed method goes beyond 2D data augmentation in the image domain and introduces new augmentation capabilities in generating arbitrarily out-of-plane rotated and relighted high quality face images.

The generated face images are expected to preserve the 3D landmark locations, identity, and the visual attributes of the input face.
So these augmented face images can be used for training various face understanding models. \figref{fig:teaser_fig_aug} illustrates some exemplar 3D augmented face images from the proposed method. 

We experiment with data augmentation using this pipeline on three major face understanding tasks: 1) face landmark localization; 2) face attribute classification; 3) face recognition. Aside from outperforming the state-of-the-art models, noticeable improvement in robustness is observed for each independent model. 
When chaining the models trained with the proposed data augmentation techniques, an end-to-end face understanding system can be built with strong robustness against pose and lighting changes. To further demonstrate the superiority of the proposed 3D modeling based data augmentation for face understanding, we conduct a comparison study of using other 2D augmentation techniques such as 2D similarity transforms.

The main contributions of this work are summarized as follows: 1) we propose a unified data augmentation pipeline based on 3D face modeling for various face understanding tasks; 2) we demonstrate face understanding models trained with the proposed data augmentation can achieve state-of-the-art accuracy with better robustness against pose and lighting changes; 3) we provide the first comprehensive and in-depth breakdown analysis, for instance, against face pose groups, on the effectiveness of data augmentation in enhancing the robustness of face understanding models; 4) we provide task-specific 3D data augmentation design strategies that are crucial to achieve significant performance improvements for different face understanding tasks.

The paper is organized as the following: \secref{sec:related_works} gives an overview of related works in robust face understanding and the general data augmentation techniques. \secref{sec:reconstruction} then details the proposed 3D data augmentation method. \secref{sec:3D_to_face_analysis} describes how to design specific augmentation strategies for the different face understanding tasks. \secref{sec:results} shows the experiments demonstrating the effectiveness of the proposed 3D data augmentation in improving robustness of face understanding models and achieving state-of-the-art (SoTA) results.

\begin{figure*}[tbh]
  \includegraphics[width=\linewidth]{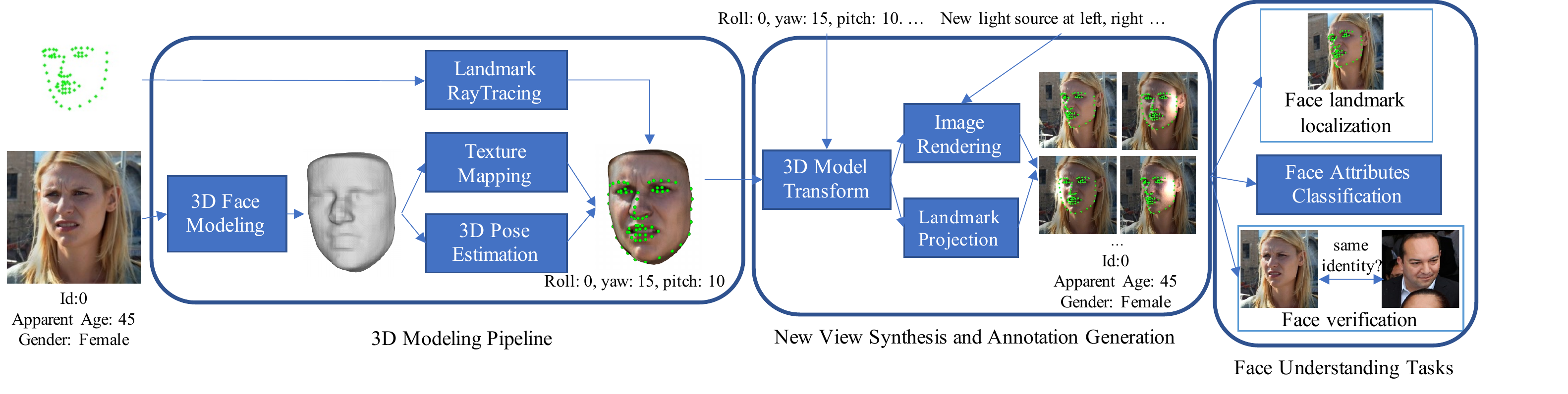}
  \caption{\label{fig:overview} Overview of the proposed 3D data augmentation method for robust face understanding. It consist of three stages: (1) A 3D modeling stage (2) New view synthesis and annotations propagation for synthesized images. (3) Applying the 3D synthesized images to various face understanding  tasks. Start with a single in-the-wild face image and its associated annotations. We first estimate its 3D face shape together with rigid-body 6dof pose and texture-map it. Second, we synthesize images with new viewpoints and new lighting conditions to form a 3D augmented data bank with high quality annotation propagated. This augmented data bank will then be applied to a comprehensive set of face understanding  tasks: face landmark localization, face attributes classification and face recognition.}
\end{figure*}

\section{Related Works\label{sec:related_works}}

In our work, we aim at robust face landmark localization, face attribute classification, and face recognition. A complete survey on the state-of-art algorithms for all these tasks is out of the scope of the work. Here, we mainly focus on literature that aim to improve the performance and robustness of face understanding  models through data augmentation and generative model.

\paragraph{Data Augmentation for Deep Learning Models} 
Data augmentation is a central topic in deep learning based methods~\cite{perez2017effectiveness}, which aims to mitigate the scarcity of training data in some aspects by using different techniques to synthesize new training data with free annotations. In~\cite{krizhevsky2012alexnet,szegedy2015googlenet,he2016renset}, simple similarity transformation and multi-scale cropping are used to strengthen object classification models' shift and scale invariance. In~\cite{wang2017afrcnn}, the data augmentation is done through purposely hiding some part of input images during training. However, the majority of these data augmentation techniques only operate on the 2D image domain and can not help with the target of this work, which is to increase the robustness of face understanding models against 6 dof pose and lighting variation. On the other hand, our method is able to close the data gap in these two dimensions that are impossible with general 2D data augmentation, through out-of-plane 3D rotation augmentation along the yaw and pitch axis of the input face image and illumination augmentation via relighting of the face image using normal / depth information of the 3D face geometry estimated.

\paragraph{Generative Adversarial Model for Face Understanding} 
In~\cite{antoniou2017data}, generative models, such as GAN~\cite{goodfellow2014generative}, are used to synthesize new training data with controlled attributes. DR-GAN \cite{tran2017disentangled} aims to dis-entangle pose and identity information by combining representation learning with adversarial generative learning. CAPG-GAN \cite{hu2018pose} uses a face landmark mask as pose-guidance in generating rotated face images. LB-GAN \cite{LB-GAN} uses a two-stage training approach to first frontalize an input face and then rotate it with a given pose-code. Zhang et al. \cite{zhang2018generative} adopts spatial attention for adversarial facial attributes editing and use the attributes-modified images as data augmentation for face recognition. FF-GAN \cite{yin2017towards} uses 3D Morphable Model \cite{Romdhani2005Estimating3S} parameters in a conditional-GAN framework to generate frontal face images for face recognition. Though our work follows a generative process, different from the GAN methods, we use a white-box approach where the 3D modeling and rendering are both fully configurable. In addition, we have three advantages: 1) we do not need image pairs that display large pose / illumination difference in training the synthesis pipeline. In fact, we do not need training at all. Starting with a pre-trained in-the-wild single-image 3D face modeling method, we can generate images in arbitrary viewpoints and lighting. Whereas, the variation of synthesis is limited by the diversity of training pairs that are compulsory in \cite{hu2018pose, yin2017towards}. 2) our 3D data augmentation is able to work on images from any domain, either in-the-wild or controlled environments, while the majority of GAN based methods have to rely on their training domain for an effective synthesis. Also, our method achieves state-of-art results not only on face recognition but also on face landmark localization and face attributes classification.

\paragraph{Data Augmentation with 3D Generative Model}
Masi et al. \cite{MasiEtAl:MillionsFaces:ECCV:2016} uses a collection of fixed 3D face geometry, with image-dependent texture maps for augmentation. Kim et al. \cite{KimEtAl:3DFaceIdentification:IJCB:2017} directly augment a 3D dataset by varying facial expressions from original 3D scans, generating novel images at multiple camera poses. In \cite{Banerjee2018FastFI}, face images are synthesized by querying a database of predefined 3D shapes and textures generating new views of both real and synthetic identities. \cite{crispell2017dataset, dataset-augmentation} both adopt a sparsely fitted 3D model from 2D face landmarks for separate pose and illumination augmentation. In \cite{DBLP:journals/corr/ZhuLLSL15}, a synthesized training set with profile views created using a multi-feature fitted 3D morphable model \cite{Romdhani2005Estimating3S} is used to improve face landmark localization. Zhao et al. \cite{zhao20183d} first generates pose synthesized images using a 3D face model fitted on sparse face landmarks and then employs a dual-agent GAN to refine synthesis quality. Our work differs from them in several aspects: 1) Our pipeline utilizes a learning based image-dependent face 3D modeling method which is more geometrically accurate than generic / sparsely fitted 3D models and leads to more effective data augmentation with higher performance and more robustness in face understanding  tasks. We demonstrate this in detail in section \ref{sec:compare_3d_aug} of the experiments. 2) Ours is able to apply simultaneously pose and illumination variation on the same arbitrary-viewpoint input image. 3) We are the first to show detailed analysis on applying 3D-based augmentation to a comprehensive set of components in the face understanding  pipeline, with fine-grained breakdown analysis, such as against pose, validating the improved robustness, providing a systematic evaluation of 3D-aided face understanding  through augmentation. 4) At the same time, we achieve state-of-art results on all face understanding tasks involved.

\section{3D Data Augmentation \label{sec:reconstruction}}
The proposed data augmentation technique aims to relieving the difficulty of training face understanding models with images from various poses and lighting.
We apply the paradigm of reconstructing and rendering, with the pipeline illustrated in \figref{fig:overview}. 
The input is a training set of 2D face images.
We reconstruct a 3D face model from any one of these 2D images, with the image pixels (textures) back-projected to the 3D surface. 
Then a random viewpoint is sampled and the 3D model is transformed to the new viewpoint and rendered with an optional new lighting condition. The rendered images are merged into the input set. 
In this process, the important information of interest on the face, such as shape, attributes, and identity, is expected to be preserved, so that the annotation on the original 2D image can be safely transferred to the newly synthesized image. 
The result of this pipeline is a significantly enlarged dataset, with better diversities in face poses and illumination, leading to 
better robustness against pose and lighting variance for our trained models. 

\paragraph{High quality data augmentation} We demand a high quality data augmentation technique to have two major properties.
The first is \emph{diversity}, where the augmented data is ought to be diverse to cover the variability in real-world scenarios. 
The second is \emph{fidelity}, where the augmentation technique itself should not introduce unrealistic artifacts that hurt the model's overall accuracy in recognizing real face images.
In the proposed approach, the \emph{diversity} is achieved because we are able to render a reconstructed 3D face model in any viewpoint with arbitrary new illumination condition.
In the remaining of the section, we will describe our 3D modeling and view synthesis pipeline to generate high \emph{fidelity} augmented data.

\subsection{3D Face Modeling\label{sec:3d_face_recon}}

The first component in the pipeline is 3D face modeling, which estimates the 3D face geometry using a ~\emph{single} face image.
Though any 3D modeling method can be integrated seamlessly, we select the volumetric shape regression method (VRN) \cite{JacksonEtAl:FaceRecon:ICCV:2017} based on its state-of-art reconstruction accuracy. 
% From each input image, the VRN model produces an image-dependent 3D occupancy volume, which are then converted to a mesh with marching-cubes \cite{LorensenAndCline:MarchingCubes:1987}. 
% This work proposes a basic architecture, the \emph{Volumetric Regression Network} (VRN), consisting of two stacked hourglass networks, with skip connections between the encoding and decoding layers.
% Several variants of the base VRN network are also proposed in their work. A variant of the basic architecture, named VRN-Guided, precedes the original VRN block with another hour-glass network, trained for regressing heatmaps of iBUG landmarks \cite{SagonasEtAl:iBUG:CVPRWorkshop:2013}. The output of this first network is concatenated with the original input image, and fed into the original VRN network (with allowances made for the new input type). They also propose a VRN-Multitask for regression of both the 3D facial volume and a set of sparse face landmarks.
% Note there are multiple variants of the base VRN network in ~\cite{JacksonEtAl:FaceRecon:ICCV:2017}. For the simplicity, we use the baseline unguided version. 
%  This method creates a unique 3D face model for each input image and thus is person dependent.
% A typical mesh output of our image-dependent 3D reconstruction is shown in \figref{fig:reconstruction}. 
Another alternative is to use a generic parametric 3D face model for all face images, such as in \cite{bfm09}. %This method is described in \cite{MasiEtAl:MillionsFaces:ECCV:2016} for the task of face identification.
% This method will ignore the inter-person variation in the 3D model space and is thus person independent. % comment: this is removed to avoid confusion, as the motivation of the paper is not to study the impact of reconstruction quality to the downstream face analysis tasks. If we want to list that as major contribution, then we should compare a broad range of 3D face reconstruction methods (VRN, 3DMM, self-supervised, generic 3D, or a ball) in improving the downstream face analysis tasks. This can be mentioned as future work.
We show experiments with 3D augmentation using both 3D modeling methods in \secref{sec:compare_3d_aug} and study the impact of different 3D modeling methods to face understanding performance.

\paragraph{3D Pose Estimation\label{sec:symmetry}}
Acquiring the 6dof pose of the 3D face w.r.t the input image's viewpoint is important for high ~\emph{fidelity} data augmentation. When generating pose augmentation, we use it to avoid showing self-occluded regions of the 3D face which has no direct texture from the image. Obtaining the pose of the 3D face depends on the underlying 3D modeling method. For parametric based methods, the 6dof pose parameters are usually directly regressed together with the shape parameters \cite{DBLP:journals/corr/TranHMM16, DBLP:journals/corr/abs-1712-02859, chang2017faceposenet}.
For other methods without direct regression such as volumetric based ones \cite{JacksonEtAl:FaceRecon:ICCV:2017, Xing2018ASB}, we follow \cite{Xing2018ASB} for 6dof pose estimation utilizing the bilateral symmetry property of the 3D face. 
% for illumination augmentation, it is necessary to get the 6dof pose of the 3D face to relight the image in the original viewpoint.
% Once the 6dof pose is obtained, it then enables augmented rendering with arbitrary viewpoints and relighting.

% rather, for which the 3D face shape estimated is in the camera coordinate system,

% The bilateral symmetry plane and back plane are extracted, which define the 6dof pose of the 3D shape through eigen-value decomposition on the face mesh vertices. 

\paragraph{Texture Mapping\label{sec:texture_map}}
% {\color{red} TODO: reduce to give more space to experiments (three facial tasks) and ablation study (comparison between generic shape and reconstructed shape)}

% \begin{figure}
%   \begin{tighttabular}{cc}
%         \includegraphics[width=0.24\columnwidth]{./Figures/O3} &\includegraphics[width=0.75\columnwidth]{./Figures/pictorial_ray_tracing} \\
%         (a) & (b)
%   \end{tighttabular}
%   \caption{\label{fig:texture_mapping} (a) Texture-mapped 3D mesh model overlaid on top of original 2D image and (b) landmark ray tracing on the volumetric shape regression output. Note that the rendered model is aligned against the original image. {\color{red} Change the texture mapping and ray-tracing plot here}}
% \end{figure}

% \begin{figure}
% \begin{subfigure}[b]{0.24\columnwidth}
%     \includegraphics[width=1.0\columnwidth]{./Figures/O3}
%     \caption{\label{fig:texture_mapping}}
% \end{subfigure}
% \begin{subfigure}[b]{0.75\columnwidth}
%     \includegraphics[width=1.0\columnwidth]{./Figures/pictorial_ray_tracing}
%     \vspace{-3em}
%     \caption{\label{fig:landmark_raytracing}}
% \end{subfigure}
% \caption{(a) Texture-mapped 3D mesh model overlaid on top of the original 2D image and (b) landmark ray tracing on the volumetric shape regression output. Note that the rendered model is aligned against the original image. {\color{red} Change the texture mapping and ray-tracing plot here}}
% \end{figure}
To achieve high \emph{fidelity}, we must preserve the color information on the original input 2D face. Consequently, instead of regressing the RGB values for the mesh vertices / faces or a 2D UV-coordinate map, we directly map the pixel information on the original 2D face image onto the 3D mesh vertices. Details of the process is described in \secref{sec:texture_map_appendix} of the appendix.
%as represented in Figure \ref{fig:texture_mapping}.

\subsection{High Quality New View Synthesis\label{sec:viewpoints}}

With the 3D face model and texture mapping ready, it is now possible to render realistic face images from the 3D models in any new arbitrary viewpoints and lighting conditions.
For 3D pose augmentation, we produce new views of the input face image via the rotation of the 3D face shape around the $x$ (pitch) and $y$ (yaw) axis. As the quality of realist rendering is crucial for high \emph{fidelity} data augmentation, we constrain the rotation angle around $y$ (yaw) not to expose self-occluded regions of the mesh. Also, we avoid large $x$ (pitch) axis rotations which will merely show the forehead or the chin of the face in the rendered image. For illumination augmentation, we place four additional light sources, at the locations of bottom, top, left and right, to the 3D face shape and randomly activate one of them during rendering. Details of the rendering set-up and view synthesis rigid-rotation constraints can be found in \secref{sec:rendering_setup_appendix} and \secref{sec:rigid_control_appendix} of the appendix. A typical result of this procedure is shown in \figref{fig:landmarks_68}~\ref{fig:render_illumination}.
\section{Applying 3D Augmentation to Face Understanding\label{sec:3D_to_face_analysis}}

% \subsection{High Quality Annotation for Face Analysis \label{sec:3D_to_face_analysis}}
A typical face understanding system \cite{Cao18, wang2018devil, wen2016discriminative, guo2016msceleb} has several modules. An input image is first examined by a face detector to localize face bounding boxes. Then face landmarks are localized for each detected face. The face crops are then normalized via 2D similarity transform based on landmark locations. Finally, face attribute classifiers and face identification models are applied on the normalized face images to extract attributes and identity information.
We now describe the crucial steps in applying the proposed 3D data augmentation method to the training datasets of three critical tasks in face understanding, \emph{i.e.} face landmark localization, face attribute classification, and face identity recognition. 
The goal of applying the proposed data augmentation pipeline is to improve the robustness of the corresponding models against strong pose and illumination variations. 

\subsection{Face Landmark Localization\label{sec:apply_to_landmark}}
We focus on the 3D landmark localization problem where the regressed 2D landmark points are projections of 3D landmark points onto the 2D image.
The training datasets for this task usually consist of images with landmarks annotated by human, which is a laborious work, limiting the availability of large-scale datasets. 
Besides, faces in profile views (large yaw rotation) and faces under challenging illumination condition are more difficult to annotate, leading to increasing annotation noise.
Consequently, models learned on these datasets tend to fail for non-frontal faces and faces with unusual lighting.
By applying the proposed pipeline, we can augment the training data with much more diverse pose and lighting distributions and accurate landmark annotation for non-frontal views. 
We design the following steps to generate high-quality ground-truth landmark locations with visibility information at new views.

\paragraph{Determining 3D Landmark Locations\label{sec:ray_tracing}}
We need accurate 3D landmark locations on the 3D mesh when projecting to 2D locations in new views. For a parametric 3D model, the 3D landmark locations are determined through a fixed topology, as in \cite{bfm09}. For non-parametric models, the 3D landmarks are generated by ray-tracing the 2D locations in the original input image onto the 3D mesh.

\begin{figure}[tbh]
\setlength{\tabcolsep}{0pt}
\begin{subfigure}[t]{0.98\linewidth}
	\tabcolsep=0pt
\begin{tabu} to \textwidth {X[0.5,c,m]*7{X[1,c,m]}}
	& original & $+20^\circ$ & $+30^\circ$ & $+40^\circ$ & $+50^\circ$ & $+60^\circ$ & $+70^\circ$\\ 
	$-20^\circ$ &
	\includegraphics[width=0.125\columnwidth]{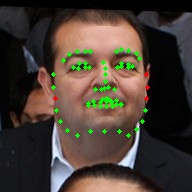}&
	\includegraphics[width=0.125\columnwidth]{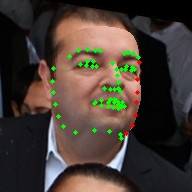}&
	\includegraphics[width=0.125\columnwidth]{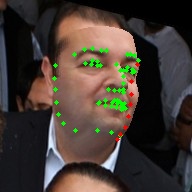}&
	\includegraphics[width=0.125\columnwidth]{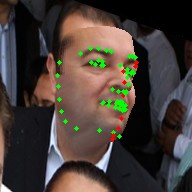}&
	\includegraphics[width=0.125\columnwidth]{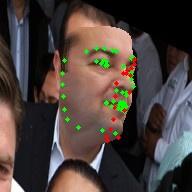}&
	\includegraphics[width=0.125\columnwidth]{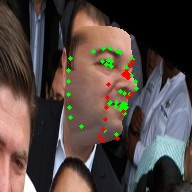}&
	\includegraphics[width=0.125\columnwidth]{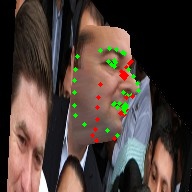}\\
	\rotatebox[origin=c]{90}{original} &
	\includegraphics[width=0.125\columnwidth]{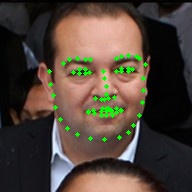}&
	\includegraphics[width=0.125\columnwidth]{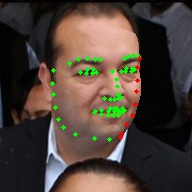}&
	\includegraphics[width=0.125\columnwidth]{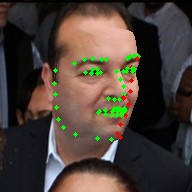}&
	\includegraphics[width=0.125\columnwidth]{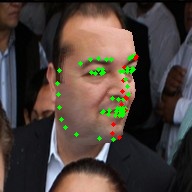}&
	\includegraphics[width=0.125\columnwidth]{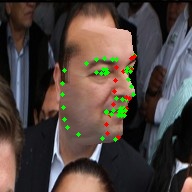}&
	\includegraphics[width=0.125\columnwidth]{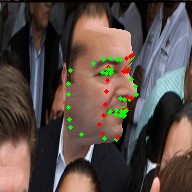}&
	\includegraphics[width=0.125\columnwidth]{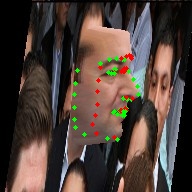}\\
	$+20^\circ$ &
	\includegraphics[width=0.125\columnwidth]{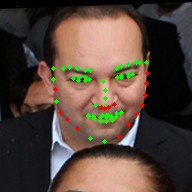}&
	\includegraphics[width=0.125\columnwidth]{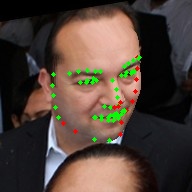}&
	\includegraphics[width=0.125\columnwidth]{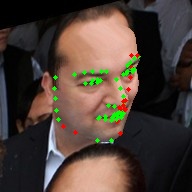}&
	\includegraphics[width=0.125\columnwidth]{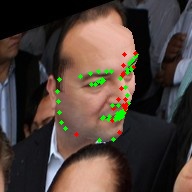}&
	\includegraphics[width=0.125\columnwidth]{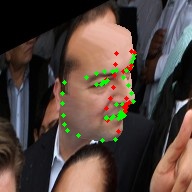}&
	\includegraphics[width=0.125\columnwidth]{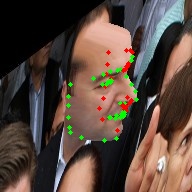}&
	\includegraphics[width=0.125\columnwidth]{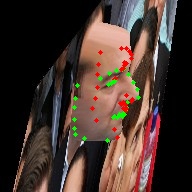}\\
\end{tabu}
	\caption{\label{fig:landmarks_68} New views with propagated face landmarks over-laid.}
\end{subfigure}
\begin{subfigure}[t]{0.98\linewidth}
	\tabcolsep=0pt
\begin{tabu} to \textwidth {X[0.5,c,m]*7{X[1,c,m]}}
	& original & $+5^\circ$ & $+10^\circ$ & $+15^\circ$ & $+20^\circ$ & $+40^\circ$ & $+60^\circ$\\ 
	$-20^\circ$ &
	\includegraphics[width=0.125\columnwidth]{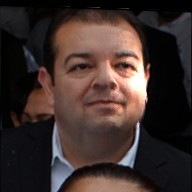}&
	\includegraphics[width=0.125\columnwidth]{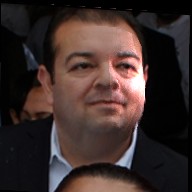}&
	\includegraphics[width=0.125\columnwidth]{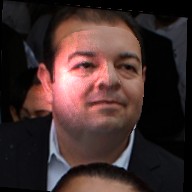}&
	\includegraphics[width=0.125\columnwidth]{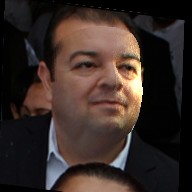}&
	\includegraphics[width=0.125\columnwidth]{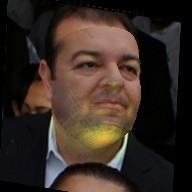}&
	\includegraphics[width=0.125\columnwidth]{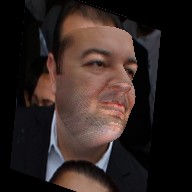}&
	\includegraphics[width=0.125\columnwidth]{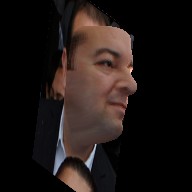}\\
	\rotatebox[origin=c]{90}{original} &
	\includegraphics[width=0.125\columnwidth]{./Figures/0000000002/0000000002_aug_0_0}&
	\includegraphics[width=0.125\columnwidth]{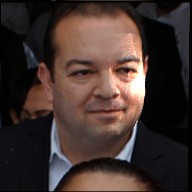}&
	\includegraphics[width=0.125\columnwidth]{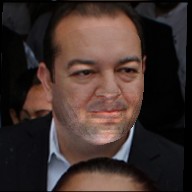}&
	\includegraphics[width=0.125\columnwidth]{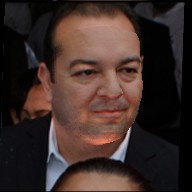}&
	\includegraphics[width=0.125\columnwidth]{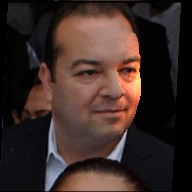}&
	\includegraphics[width=0.125\columnwidth]{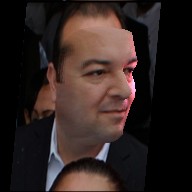}&
	\includegraphics[width=0.125\columnwidth]{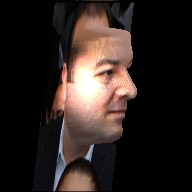}\\
	$+20^\circ$ &
	\includegraphics[width=0.125\columnwidth]{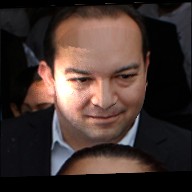}&
	\includegraphics[width=0.125\columnwidth]{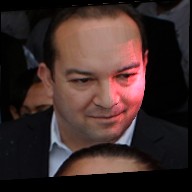}&
	\includegraphics[width=0.125\columnwidth]{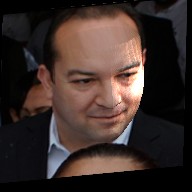}&
	\includegraphics[width=0.125\columnwidth]{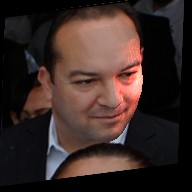}&
	\includegraphics[width=0.125\columnwidth]{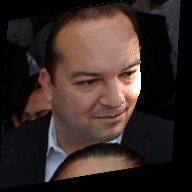}&
	\includegraphics[width=0.125\columnwidth]{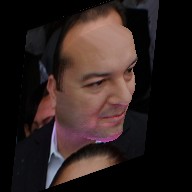}&
	\includegraphics[width=0.125\columnwidth]{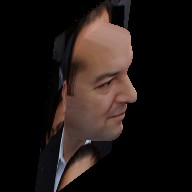}\\
\end{tabu}
	\caption{\label{fig:render_illumination} New views with illumination augmentations.}
\end{subfigure}
\caption{(a) 3D pose augmented images in new views with the propagated 68 landmark locations over-laid. Landmarks in red are occluded; green means visible. (b) Synthesized views with illumination augmentation of additional light sources. In both figures, columns and rows represent yaw and pitch rotation respectively.}
\end{figure}

\paragraph{Landmark Visibility from New Views}
It is necessary to mark the visibility of a 3D landmark in synthesized images after 3D pose rotation for training the augmented landmark models.
This is achieved by computing the 3D euclidean distance on the mesh coordinate space between a 3D landmark and the intersection of a ray that goes from the camera origin to that 3D landmark location with the mesh. A distance threshold is used to determine such visibility when projected to new views. Examples are shown in \figref{fig:landmarks_68}.

\paragraph{Augmentation Strategy} When training pose augmented landmark models, we augment each near-frontal image in the training data to random views of $\pm20^{\circ}\text{\!\!,}$ $\pm40^{\circ}$ in yaw. For illumination augmentation, each near-frontal image is augmented with a new lighting setup described in \secref{sec:viewpoints}.

\subsection{Face Recognition}
Face recognition aims to recognize the identity of an input face. To perform data augmentation, we directly propagate the original identity information to the newly synthesized images. 
Since face recognition models operate on aligned face images, we follow the standard 5-point alignment process described in \cite{Cao18}. During the training, we adopt the same illumination and pose augmentation strategy as \secref{sec:apply_to_landmark} on near-frontal images.
\paragraph{Dealing with identity long-tail distribution} As the training data for face recognition is extremely long-tail regarding the number of images per identity, how the synthesized new views are added into the training data is crucial.  In addition, we experiment with two methods of sampling the newly augmented images during training. The first method is random sampling where the 3D augmented images will be sampled randomly according to a probability. For both illumination and pose augmented images, we keep the percentage of the synthetic to real below a threshold of 1/2. The second method will first group the training images into their identity labels. Then, the pose distribution of each identity is calculated via placing its images into a discrete set of pose groups. \figref{fig:sample_entropy} shows the yaw pose distribution of a sample identity in the TrillionPairs\cite{TrillionPairs} training set. Subsequently, the entropy statistic of pose distribution is calculated for each identity with the equation below. 
\begin{equation}
    E(identity) = \sum_{i=1}^{n} -p_{i}log_{p_{i}} \label{eq:yaw_entropy}
\end{equation}
where $n$ is the number of pose groups and $p_{i}$ is the density of group $i$ for that identity. Lastly, we use an entropy cut-off threshold such that pose augmentation will only be performed on images of identities having yaw entropy ~\emph{smaller} than this threshold. The hypothesis is that if an identity already contains rich pose variation in its training images, we do not need to generated more pose augmented images for it. In \secref{sec:ablation_experiment}, we show that the ratio of synthetic vs real images and the sampling scheme have significant impacts on the verification and identification results.

% \figref{fig:cross_entropy} shows the sample cross-entropy distribution of a particular face recognition dataset - TrillionPairs \cite{TrillionPairs}.

\subsection{Face Attributes Classification}
For face attributes classification, we consider two common attributes: age and gender. We directly transfer the age and gender annotation from the original input image to the newly synthesized views as facial pose or illumination condition variation would not alter these demographic attributes. %As such, the transferred attributes will have the same exact quality of input face images. 
During training, we augment each frontal face image in the dataset to random views of $\pm10^{\circ}$, $\pm20^{\circ}\text{\!\!,}$ $\pm40^{\circ}$, $\pm60^{\circ}$ in yaw and $\pm20^{\circ}$ in pitch. For illumination augmentation, similar to face landmark localization and face recognition, we randomly select the four additional light sources to change the lighting for each synthesized image.

%%% Local Variables:
%%% mode: latex
%%% TeX-master: "data_augmentation_for_face_recognition"
%%% End:

\section{Experiments\label{sec:results}}

We experiment applying the proposed data augmentation to three face understanding tasks: face landmark localization, face attribute classification, and face identity recognition, on two dimensions: pose and illumination augmentation, respectively.

\subsection{Face Landmark Localization}
We train our face landmark localization models on LS3D-W~\cite{bulat2017far} dataset. LS3D-W \cite{bulat2017far} is a large and challenging 68-point 3D landmark dataset that unifies the majority of existing face landmark datasets to date. We split it into a $90\%$ - $10\%$ training/testing set. For evaluation, we use the AFLW2000-3D \cite{DBLP:journals/corr/ZhuLLSL15} benchmark and the testing split of LS3D-W. The evaluation metric is the Normalized Mean Error \cite{bulat2017far} defined as the 2D euclidean distance between the ground truth landmark locations and the predicted locations normalized by the detected bounding box size.
The pipeline in \figref{fig:overview} %and described in detail in sections \secref{sec:reconstruction} through \sefref{sec:bag_of_tricks} 
was applied to the LS3D-W \cite{bulat2017far} dataset. We exclude the LS3D-W Balanced and AFLW2000-3D re-annotated set in training and 3D augmentation to avoid train-test overlap.
% 229,505 images were fed into the pipeline, excluding the LS3D-W Balanced set. From the ray traced 3D landmarks and the symmetry measurement defined in \secref{sec:symmetry}, 133943 were accepted and used to produce the final rendering results and projected landmark annotations for the new views, $t_{\text{3D}}$ of 5.0 is chosen for post-filtering from the histogram of symmetry measurement on this dataset. Examples of rendered images from different pitch and yaw pose augmentation are shown in \figref{fig:landmarks_68}
Our landmark regression network uses ResNet-34~\cite{he2016renset} architecture and directly regresses the 2D coordinates of the $68$ landmarks. We report performances of the model trained with and without the proposed 3D data augmentation on dimension of pose and lighting.

\begin{figure*}[tbh]
  \begin{subfigure}[t]{0.245\linewidth}
        \includegraphics[width=1.\columnwidth]{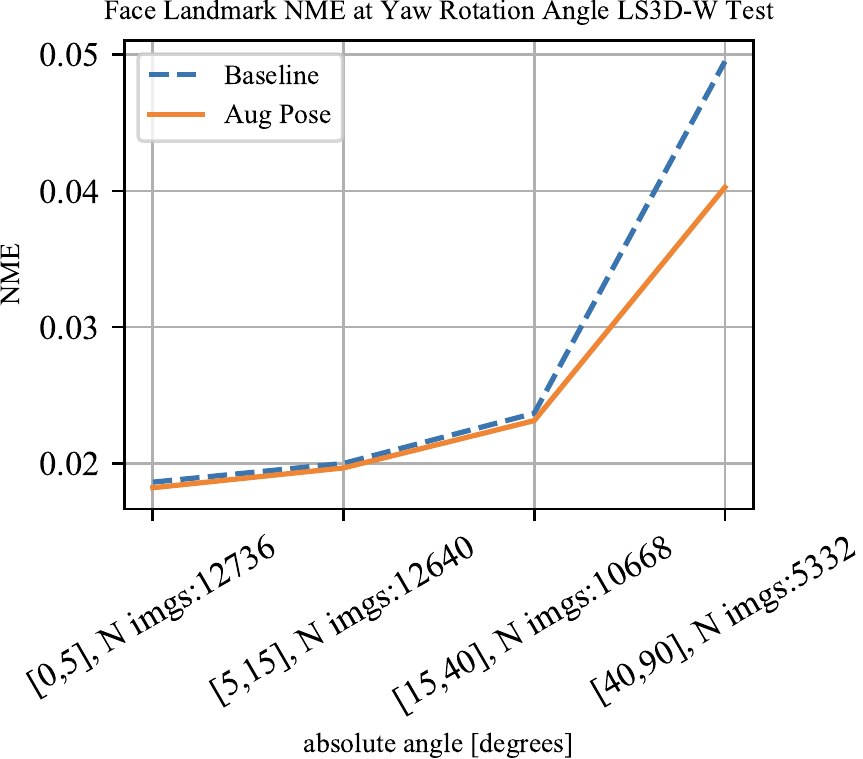}
        \caption{}\label{fig:pose_plot_lmk}
    \end{subfigure}
    \begin{subfigure}[t]{0.245\linewidth}
        \includegraphics[width=1.0\columnwidth]{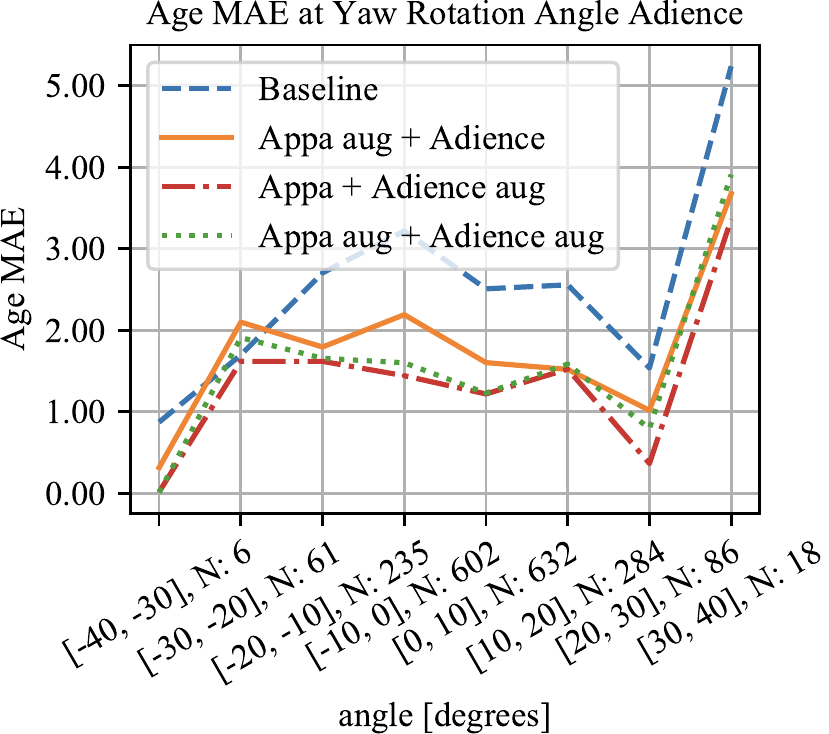}
        \caption{}\label{fig:pose_plot_age}
    \end{subfigure}
    \begin{subfigure}[t]{0.245\linewidth}
        \includegraphics[width=1.0\columnwidth]{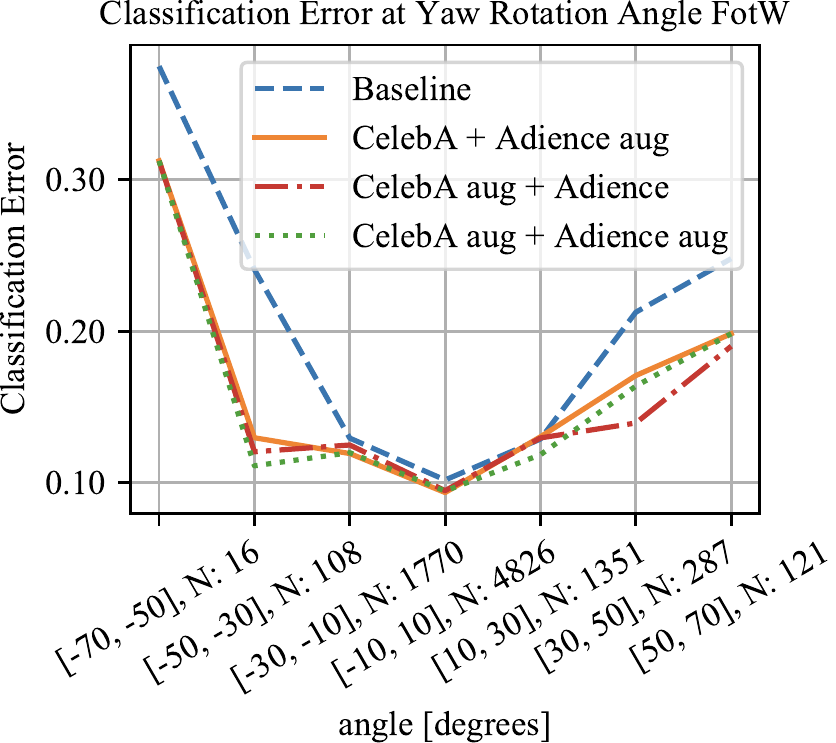}
        \caption{}\label{fig:pose_plot_gender}
    \end{subfigure}
    \begin{subfigure}[t]{0.245\linewidth}
        \includegraphics[width=1.0\columnwidth]{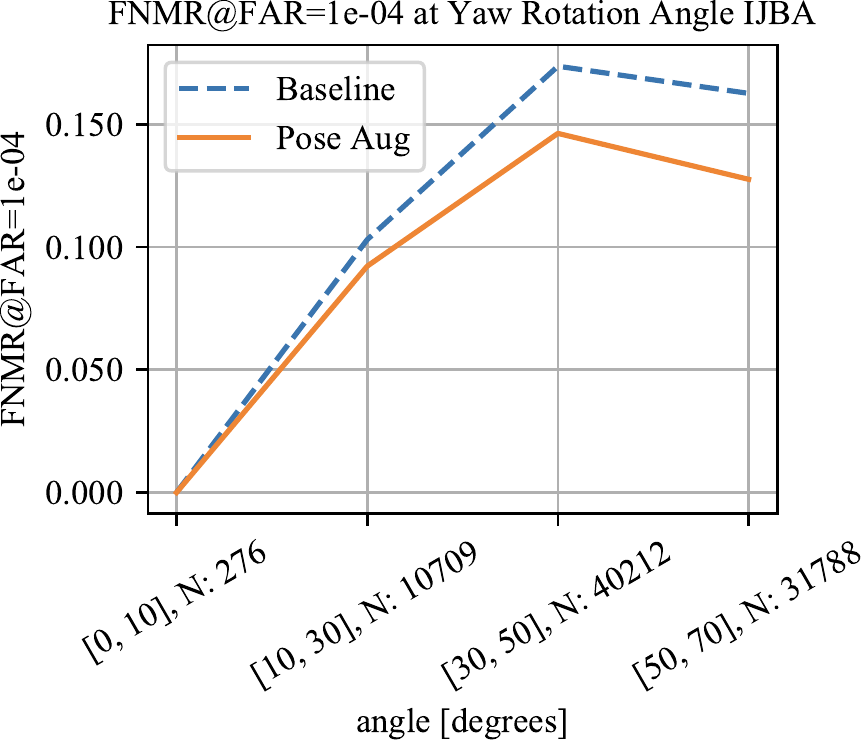}
        \caption{}\label{fig:ijba_plot_compare_yaw}
    \end{subfigure}
  \caption{\label{fig:angle_analysis} Fine-grained comparison of baseline models and models trained with 3D data augmentation against pose as a function of yaw angle of input images. Here, (a) face landmark localization error (NME, lower better, plotted against absolute yaw angle) on LS3D-W \cite{bulat2017far} testset, (b) apparent age prediction error (MAE, lower better) on Adience \cite{eidinger2014age},  (c) apparent gender classification error (lower, better) on FotW \cite{escalera2016chalearn} and (d) IJBA \cite{ijba} template verification error (lower, better, plotted against absolute yaw angle). Improvement is significant for all pose groups under all tasks, especially at large viewpoints}
\end{figure*}

\tabref{tbl:compare_AFLW2000_3D} shows the results on AFLW2000-3D. Overall, models trained with the proposed augmentation achieve lower NME. We also provide a breakdown analysis to assess the model robustness to pose variation. \figref{fig:pose_plot_lmk} shows the NME on LS3D-W test split against the yaw pose angle of test images. It shows that the landmark model trained with 3D pose augmentation significantly outperforms the baseline model with much lower NME on images with large yaw angles, demonstrating the benefit of the proposed method in improving model robustness against pose variation.

% \begin{table}[bth]
% \begin{footnotesize}
% \setlength{\tabcolsep}{1pt}
% \centering
%     \begin{tabular}{>{\centering\arraybackslash}p{5.0cm}>{\centering\arraybackslash}p{2.3cm}}%{|>{\centering\arraybackslash}p{4.0cm}||>{\centering\arraybackslash}p{2.3cm}|>{\centering\arraybackslash}p{1.1cm}|}
%     	Model & LS3D-W Test\\ [0.5ex] 
%     	\hline\hline
%     	Baseline & 2.78 \\ 
%     	\hline
%     	Augmented $\pm40^{\circ},\pm20^{\circ}$ yaw (Ours) & \textbf{2.58} \\
%     	\hline
%     \end{tabular}
%     \caption{\label{tbl:landmark_aug_compare} NME error (\%) on LS3D-W \cite{bulat2017far} Test set.}
% \end{footnotesize}
% \end{table}

\begin{table}[tbh]
\begin{center}
% \vspace{-0.9em}
\begin{footnotesize}
\setlength{\tabcolsep}{1pt}
\centering
\begin{minipage}[t]{1.0\linewidth}
\begin{adjustbox}{max width=\linewidth}
    \begin{tabular}{>{\centering\arraybackslash}p{0.5cm}>{\centering\arraybackslash}p{1.5cm}>{\centering\arraybackslash}p{1.5cm}>{\centering\arraybackslash}p{1.5cm}>{\centering\arraybackslash}p{1.6cm}>{\centering\arraybackslash}p{1.6cm}|>{\centering\arraybackslash}p{1.0cm}>{\centering\arraybackslash}p{1.15cm}>{\centering\arraybackslash}p{1.15cm}}
    	& RCPR \cite{PCPR} & ESR \cite{ESR} & SDM \cite{SDM} & 3DDFA \cite{DBLP:journals/corr/ZhuLLSL15} & 3DSTN \cite{3DSTN} & Baseline & A.Illum (Ours)  & A.Pose (Ours)\\[0.5ex] 
    	\hline
    	\hline
    	Mean&  7.80 & 7.99 & 6.12 & 5.42 & 4.49 &3.85 & 3.72 &  \textbf{3.51}\\ 
    	\hline
    	Std & 4.74 & 4.19 & 3.21 & 2.21 & \textbf{1.42} & 2.34 & 2.29 & 3.28 \\ 
    	\hline
    \end{tabular}
    % \vspace{-1em}
    \end{adjustbox}
\end{minipage}
\caption{\label{tbl:compare_AFLW2000_3D} Landmark localization error (NME, \%) results on AFLW2000-3D benchmark. A.Illum and A.Pose refer to illumination and pose augmentation respectively}
\end{footnotesize}
\end{center}
\end{table}

% one table for all age experiments
% \begin{table}[tbh]
% \begin{footnotesize}
% \setlength{\tabcolsep}{1pt}
% \centering
%     \begin{tabular}{>{\centering\arraybackslash}p{0.9cm}>{\centering\arraybackslash}p{0.9cm}>{\centering\arraybackslash}p{0.8cm}|>{\centering\arraybackslash}p{0.9cm}>{\centering\arraybackslash}p{0.9cm}>{\centering\arraybackslash}p{0.9cm}>{\centering\arraybackslash}p{0.9cm}|>{\centering\arraybackslash}p{0.95cm}>{\centering\arraybackslash}p{0.95cm}>{\centering\arraybackslash}p{0.95cm}}
%     	 & Eidinger \cite{eidinger2014age} & Levi \cite{levi2015age} & Baseline  & Appa A.Pose (Ours) & Adience A.Pose (Ours) & Both A.pose (Ours) & Appa A.Illum (Ours) & Adience A.Illum (Ours) & Both A.Illum (Ours)\\[0.5ex] 
%     	\hline
%     	\hline
%         MAE &-  & - & 2.71 & 1.80 & \textbf{1.37} & 1.47 & 2.53& 1.57& 2.00\\ 
%     	\hline
%     	Exact$\%$& 45.1 & 50.7 & 73.3  & 82.0 & \textbf{86.2} & 84.7 & 74.3& 85.2& 80.9\\ 
%     	\hline
%     	1-off$\%$ &79.5  & 84.7 & 96.6 & 96.5 & 96.9 & 96.7 & 96.2& \textbf{97.3}& 96.9\\ 
%     	\hline
%     \end{tabular}
%     \vspace{-1em}
%     \caption{\label{tbl:compare_age_classification} Age classification-accuracy($\%$) and prediction error (MAE) on Adience compared to SoTA. A.Pose and A.Illum are pose and illumination augmentation.
%     }
% \end{footnotesize}
% \end{table}

\begin{table}[tbh]
\begin{center}
\begin{footnotesize}
\setlength{\tabcolsep}{1pt}
\centering
    \begin{adjustbox}{max width=\linewidth}
            \begin{tabular}{>{\centering\arraybackslash}p{2.5cm}|>{\centering\arraybackslash}p{1.7cm}>{\centering\arraybackslash}p{1.2cm}>{\centering\arraybackslash}p{1.3cm}>{\centering\arraybackslash}p{1.1cm}}
    	 Augmentation Type & Baseline (Appa+Adience original) & Appa aug (Ours)  & Adience aug (Ours)  & Both aug (Ours) \\[0.5ex] 
    	\hline
    	\hline
    	Pose & 2.71 & 1.80 & \textbf{1.37} & 1.47 \\ 
    	\hline
    	Illumination & 2.71 & 2.53 & \textbf{1.57} &  2.00 \\ 
    	\hline
    \end{tabular}
    \end{adjustbox}
    \caption{\label{tbl:compare_age_mae} Age prediction error (MAE) on Adience \cite{levi2015age} test.}
\end{footnotesize}
\end{center}
\end{table}

\begin{table}[tbh]
\begin{center}
\begin{footnotesize}
\setlength{\tabcolsep}{1pt}
\centering
    \begin{adjustbox}{max width=\linewidth}
    \begin{tabular}{>{\centering\arraybackslash}p{0.7cm}>{\centering\arraybackslash}p{0.9cm}>{\centering\arraybackslash}p{0.8cm}|>{\centering\arraybackslash}p{1.0cm}>{\centering\arraybackslash}p{0.8cm}>{\centering\arraybackslash}p{0.8cm}>{\centering\arraybackslash}p{1.0cm}|>{\centering\arraybackslash}p{0.9cm}>{\centering\arraybackslash}p{0.9cm}>{\centering\arraybackslash}p{0.9cm}}
    	 & Eidinger \cite{eidinger2014age} & Levi \cite{levi2015age} & Baseline  & Appa A.Pose (Ours) & Adience A.Pose (Ours) & Both A.pose (Ours) & Appa A.Illum (Ours) & Adience A.Illum (Ours) & Both A.Illum (Ours)\\[0.5ex] 
    	\hline
    	\hline
    	Exact& 45.1 & 50.7 & 73.3  & 82.0 & \textbf{86.2} & 84.7 & 74.3& 85.2& 80.9\\ 
    	\hline
    	1-off &79.5  & 84.7 & 96.6 & 96.5 & 96.9 & 96.7 & 96.2& \textbf{97.3}& 96.9\\ 
    	\hline
    \end{tabular}
    \end{adjustbox}
    \caption{\label{tbl:compare_age_classification} Age classification accuracy(\%) on Adience \cite{levi2015age} test. \emph{A.Pose} and \emph{A.Illum} refer to pose and illumination augmentation. \emph{Both} refers to augmentation on both Appa and Adience training set.}
\end{footnotesize}
\end{center}
\end{table}

% \caption{\label{tbl:quantitative_age} Age prediction error (MAE) on Adience test\cite{eidinger2014age} with two types of 3D augmentation: pose and illumination.}

\begin{table}[bth]
\begin{center}
\begin{footnotesize}
\setlength{\tabcolsep}{1pt}
\centering
    \begin{minipage}[t]{1.0\linewidth}
        \begin{adjustbox}{max width=\linewidth}
                \begin{tabular}{>{\centering\arraybackslash}p{0.7cm}>{\centering\arraybackslash}p{1.4cm}>{\centering\arraybackslash}p{1.4cm}>{\centering\arraybackslash}p{1.2cm}|>{\centering\arraybackslash}p{1.1cm}>{\centering\arraybackslash}p{1.1cm}>{\centering\arraybackslash}p{1.1cm}}
                     DIF \cite{han2018heterogeneous} & MCNN AUX \cite{hand2016attributes} & Face Tracker \cite{kumar2008facetracer} & CTS CNN \cite{zhong2016face} & Baseline & A.Illum (Ours) & A.Pose (Ours)\\
                    \hline \hline
                    98 & 98 & 91 & \textbf{99}& 98.38 & \textbf{98.50} & 98.47 \\[0.5ex] 
                    \hline
                \end{tabular}
        \end{adjustbox}
    \end{minipage}
    \caption{\label{tbl:celeba_quatitative} Gender classification accuracy (\%)  on CelebA \cite{liu2015deep}}
    \end{footnotesize}
\end{center}
\end{table}

\begin{table}[bth]
\begin{center}
\begin{footnotesize}
\setlength{\tabcolsep}{1pt}
\centering
    \begin{minipage}[t]{1.0\linewidth}
        \begin{adjustbox}{max width=\linewidth}
                \begin{tabular}{>{\centering\arraybackslash}p{0.6cm}>{\centering\arraybackslash}p{0.8cm}>{\centering\arraybackslash}p{1.4cm}>{\centering\arraybackslash}p{0.8cm}>{\centering\arraybackslash}p{0.7cm}>{\centering\arraybackslash}p{1.1cm}|>{\centering\arraybackslash}p{1.1cm}>{\centering\arraybackslash}p{1cm}>{\centering\arraybackslash}p{0.9cm}}
                 DIF \cite{han2018heterogeneous} &CMP+ ETH&SMILELAB NEU&VISI. CRIM&IVA NLPR&SIAT MMLAB& Baseline & A.Illum (Ours) & A.Pose (Ours)\\[0.5ex]
                 \hline\hline
                 84.9 & 74.6 & 90.0 & 90.2 & 91.5 & \textbf{92.7} & 88.3 & \textbf{89.7} & 89.6 \\
                \hline
                \end{tabular}
        \end{adjustbox}
    \end{minipage}
    \caption{\label{tbl:fotw_quatitative} Gender classification accuracy (\%) on FotW with SoTA reported in \cite{escalera2016chalearn}}
\end{footnotesize}
\end{center}
\end{table}

\begin{table*}[tbh]
\begin{center}
\begin{footnotesize}
\setlength{\tabcolsep}{0pt}
\centering
\begin{adjustbox}{max width=\linewidth}
    \begin{tabular}{>{\centering\arraybackslash}p{3.5cm}>{\centering\arraybackslash}p{0.9cm}>{\centering\arraybackslash}p{1.0cm}>{\centering\arraybackslash}p{1.0cm}|>{\centering\arraybackslash}p{1.2cm}>{\centering\arraybackslash}p{1.0cm}>{\centering\arraybackslash}p{0.8cm}}
        Model Training & \multicolumn{3}{c}{IJBA 1:1 Verif TAR@FAR}& \multicolumn{2}{c}{1:N TPIR@FPIR} & Rank \\
        \hline
    	 & 1e-2 & 1e-3 & 1e-4 & 0.1 & 0.01 & 1 \\ [0.5ex] 
    	\hline\hline
    	vggface2 \cite{Cao18} & 0.968  & 0.921 & - & 0.946 & 0.883 & 0.982 \\ 
    	\hline
    	UMD-Face \cite{DBLP:journals/corr/abs-1809-07586} & 0.969&	0.952&	0.921 & 0.962 &	0.92 & 0.975\\ 
    	\hline
    	L2-Face \cite{ranjan2017l2} & 0.970 & 0.943 & 0.909 & 0.956 & 0.915 & 0.973 \\
    	\hline
    	MN-vc \cite{DBLP:journals/corr/abs-1807-09192} & 0.962 & 0.920 & - & - & - & -\\ 
        \hline
        DA-GAN\textsubscript{2.0} \cite{zhao20183d} & \textbf{0.989} & \textbf{0.973} & 0.946 &
        \textbf{0.982} & 0.939 & \textbf{0.990} \\
        \hline
    	Shi et al. \cite{shi2020towards} & -  & 0.963 & 0.950 & - & - & 0.975  \\ 
    	\hline
    	\hline
        Baseline*& 0.984 & 0.972 & \textbf{0.959}  & 0.975	& 0.939 & 0.984 \\
    	\hline
    	A.Illum*& 0.982 & 0.971 & 0.956 & 0.973 & 0.936 & 0.981 \\
    	\hline
    	A.Pose* & 0.984 & \textbf{0.973} &	\textbf{0.959} & 0.976 & \textbf{0.947} & 0.985 \\
    	\hline
    	A.Pose+A.Illum*&0.983 & 0.972 & 0.954 & 0.973 & 0.939& 0.984 \\
    	\hline
    \end{tabular}
    \quad
    \begin{tabular}{>{\centering\arraybackslash}p{3.0cm}>{\centering\arraybackslash}p{1.0cm}>{\centering\arraybackslash}p{1.0cm}>{\centering\arraybackslash}p{1.0cm}>{\centering\arraybackslash}p{1.0cm}|>{\centering\arraybackslash}p{1.0cm}>{\centering\arraybackslash}p{1.0cm}>{\centering\arraybackslash}p{0.8cm}}
        Model Training & \multicolumn{4}{c}{IJBC 1:1 Verif TAR@FAR}& \multicolumn{2}{c}{1:N TPIR@FPIR} & Rank \\
        \hline
        & 1e-2 & 1e-3 & 1e-4 & 1e-5 & 0.1 & 0.01 & 1 \\ [0.5ex] 
    	\hline\hline
    	vggface2 \cite{Cao18} & 0.967 & 0.927 & 0.862 & -& 0.865 & 0.763 & 0.914 \\ 
    	\hline
    	MN-vc \cite{DBLP:journals/corr/abs-1807-09192} & 0.968 & 0.927 & 0.862 & - & - & - & - \\
    	\hline
    	DCN(Divs) \cite{DBLP:journals/corr/abs-1807-11440} & 0.983& 0.947& 0.885 & - &- & -& - \\
    	\hline
    	Center Loss Features \cite{DBLP:journals/corr/abs-1809-07586} &0.953 & 0.912& 0.853 & 0.781& 0.864& 0.791& 0.917\\
    	\hline
    	UMD-Face \cite{DBLP:journals/corr/abs-1809-07586} & 0.979 &0.959& 0.925 & 0.869 & 0.9255 & 0.873 & 0.949 \\ 
    	\hline
    	Arcface \cite{deng2019arcface} & 0.9818 & \textbf{0.972}& \textbf{0.9565} & \textbf{0.9315} & - & - \\ 
    	\hline
    	\hline
        Baseline* & 0.983 &	0.970 &	0.947  & 0.888 &	0.949	 & 0.902 & 0.956 \\
    	\hline
    	A.Illum* &0.982 &0.968 &0.945 & 0.906 & 0.945 & 0.910 & 0.957\\
    	\hline
    	A.Pose* &\textbf{0.984}& \textbf{0.972}	&0.953	& 0.925 & \textbf{0.952}& \textbf{0.925} &\textbf{0.963}\\
    	\hline
    	A.Pose + A.Illum* & 0.983& 0.970&0.949 & 0.904 & 0.949 & 0.904& 0.957\\
    	\hline
    \end{tabular}
\end{adjustbox}
        \caption{\label{tbl:compare_ijba_ijbc_sota} Performance on IJBA \cite{ijba} (left) and IJBC \cite{ijbc} (right) of our approach compared to SoTA algorithms. Models are trained on the TrillionPairs \cite{TrillionPairs} dataset. A.Pose and A.Illum refer to 3D pose and illumination augmentation model training. * refers to Media-Pooling as in \cite{DBLP:journals/corr/abs-1809-07586}}
\end{footnotesize}
\end{center}
\end{table*}

\begin{table*}[tbh]
\begin{center}
\begin{footnotesize}
\setlength{\tabcolsep}{0pt}
\centering
% \begin{minipage}[t]{0.53\linewidth}
\begin{adjustbox}{max width=\linewidth}
    \begin{tabular}{>{\centering\arraybackslash}p{3.2cm}>{\centering\arraybackslash}p{1.2cm}>{\centering\arraybackslash}p{1.2cm}>{\centering\arraybackslash}p{1.2cm}>{\centering\arraybackslash}p{1.2cm}}
        Model Training & \multicolumn{4}{c}{IJBC Covariate 1:1 Verification TAR@FAR}  \\
        \hline
        & 1e-1 & 1e-2 & 1e-3 & 1e-4 \\ [0.5ex] 
    	\hline\hline
    	VGGFace2* \cite{Cao18} & 0.925&	0.86&	0.728&	0.56 \\
        MR-J(W) \cite{Mynepalli-2019-117203} & 0.95&	0.9	&0.809&	0.68 \\
    	\hline
    	\hline
    	Baseline IMDB & 0.984&	0.953	&0.890&	0.728  \\ 
    	\hline
    	Aug Pose IMDB  &\textbf{0.987}&	\textbf{0.962}	& 0.908&	\textbf{0.774} \\
    	\hline
    	Baseline TrillionPairs & 0.978&	0.951	&0.862&	0.464 \\ 
    	\hline
    	Aug Pose TrillionPairs  &0.979&	0.956&	\textbf{0.910}&	0.748 \\
    	\hline
    \end{tabular}
    \quad
    \begin{tabular}{>{\centering\arraybackslash}p{0.6cm}|>{\centering\arraybackslash}p{1.3cm}|>{\centering\arraybackslash}p{1.2cm}>{\centering\arraybackslash}p{1.2cm}>{\centering\arraybackslash}p{1.2cm}>{\centering\arraybackslash}p{1.2cm}>{\centering\arraybackslash}p{1.2cm}>{\centering\arraybackslash}p{1.2cm}}
        & &\multicolumn{5}{c}{TAR@Yaw Group} \\
        FAR & Model & [0, 10] & [10, 30] & [30, 50]& [50, 70] & [70, 90] \\ [0.5ex] 
    	\hline\hline
    	\multirow{2}{*}{1e-1}  & Baseline & 0.992 & 0.989 & 	0.986&0.978 & 	0.962 \\
                               & Aug Pose&  \textbf{0.993}&	\textbf{0.991}&	\textbf{0.989}&	\textbf{0.9786}&	\textbf{0.971}\\
                               \hline
    	 \multirow{2}{*}{1e-2} & Baseline &	0.980	&0.971&	0.954&	0.917&	0.859 \\
                               & Aug Pose& \textbf{0.982}&	\textbf{0.975}	&\textbf{0.966}&	\textbf{0.931}	&\textbf{0.886}\\
                               \hline
    	\multirow{2}{*}{1e-3}  & Baseline &0.959&	0.937	&0.883&	0.799	&0.677\\
                               & Aug Pose&\textbf{0.960}&	\textbf{0.945}	&\textbf{0.910}&	\textbf{0.831}&	\textbf{0.713} \\
                               \hline
    	\multirow{2}{*}{1e-4}  & Baseline& 	0.897&	0.837	&0.674&	0.528	&0.386 \\
                               & Aug Pose&\textbf{0.905}&	\textbf{0.867}&	\textbf{0.742}&	\textbf{0.599}	&\textbf{0.441} \\
    	\hline
    \end{tabular}
    % \vspace{-1em}
    \end{adjustbox}
    % \end{minipage}
\end{footnotesize}
    \caption{\label{tbl:compare_ijbc_covarite_sota} Performance on IJBC \cite{ijbc} 1:1 Covariate Verification benchmark (left) of models trained on IMDB \cite{wang2018devil} and TrillionPairs \cite{TrillionPairs} with and without 3D pose data augmentation. * Results of VGGFace2 other than 1e-3 is read off the plots in \cite{Mynepalli-2019-117203}; Performance breakdown (right) against the yaw angles of input image pair for models trained on IMDB \cite{wang2018devil}. The maximum yaw angle of the image pair is used as the pair angle statistic.}
\end{center}
\end{table*}

\subsection{Face Attributes Classification.}
For face attributes classification, we mainly investigate age prediction and gender classification. In this and all following tasks, we detect faces using a SSD based face detector and follow the five-point based alignment \cite{wang2018cosface}. For age prediction, we train classification-based models on the union of training splits of the Appa \cite{agustsson2017apparent} and the Adience \cite{eidinger2014age} datasets, and we test on the test split of the Adience \cite{eidinger2014age} dataset. Metric of mean absolute error (MAE) is used. For face gender classification, we train the models and generate 3D synthesized images on the training and validation splits of CelebA \cite{liu2015deep} as well as the Adience \cite{eidinger2014age} dataset, while we test on the FotW \cite{escalera2016chalearn} and CelebA test split. Standard classification accuracy is reported as the evaluation metric.
We use the ResNet-34~\cite{he2016renset} CNN as the backbone model for both tasks. 

For age prediction, results of models trained with and without pose and illumination augmentation on the Adience test split is shown in \tabref{tbl:compare_age_mae}. We further compare with the SoTA methods on this benchmark using the metric of classification accuracy in \tabref{tbl:compare_age_classification}.
For gender classification, \tabref{tbl:celeba_quatitative} shows the results on the CelebA benchmark, while \tabref{tbl:fotw_quatitative} on the FotW benchmark. In both benchmarks, the models trained with the proposed data augmentation significantly outperform the baseline model trained with real images only while achieving comparable results with the state-of-arts. \figref{fig:pose_plot_age} and \figref{fig:pose_plot_gender} demonstrates the error reduction as a function in input face image pose.

\subsection{Face Recognition}
For face recognition, we use IMDB faces \cite{wang2018devil} and TrillionPairs \cite{TrillionPairs} for training and  IJB-A \cite{ijba}/IJB-C \cite{ijbc} for evaluation. These benchmarks have two template based protocols for face verification and face identification, where each template contains a collection of face images. For face verification, we report the $1:1$ verification results in True Acceptance Rate (TAR) at different False Acceptance Rate (FAR). For face identification, we report the TPIR at different FPIR and the rank-1 search accuracy. We additionally report performances on the IJBC 1:1 Covariate image to image face verification protocol with pose breakdown analysis to validate the improved robustness of face recognition models trained with 3D data augmentation.
We use ResNet-101~\cite{he2016renset} as the backbone CNN architecture and train the face embedding network with a 256 dimensional embedding using the large margin cosine loss~\cite{wang2018cosface}.

% \footnotetext{This is read off the table in \cite{DBLP:journals/corr/abs-1809-07586}}

\tabref{tbl:compare_ijba_ijbc_sota} shows the results of our method compared with the state-of-arts on the IJBA \cite{ijba} and IJBC \cite{ijbc} benchmarks. \tabref{tbl:compare_ijbc_covarite_sota} shows results on IJBC 1:1 Covariate benchmark. On all benchmarks, our models trained with the proposed 3D augmentation achieve significantly better performance and robustness than the baselines. Also, they achieve the state-of-arts as UMD-Face \cite{DBLP:journals/corr/abs-1809-07586} and Arcface \cite{deng2019arcface} and out-perform other synthesis / augmentation methods including the GAN based ones such as DA-GAN \cite{zhao20183d}. For the full results of training with different 3D augmentation on the IMDB \cite{wang2018devil} and TrillionPairs \cite{TrillionPairs} datasets over the IJBA and IJBC benchmarks, please see \secref{sec:full_results} of the appendix.

\begin{figure}[tbh]
    \begin{subfigure}[t]{0.49\linewidth}
        \includegraphics[width=1.\columnwidth]{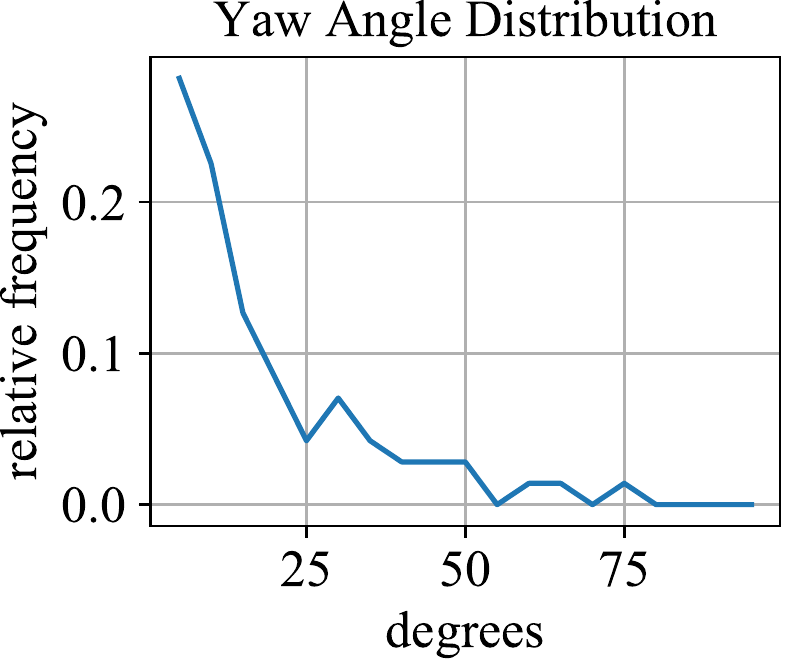}
        \caption{}\label{fig:sample_entropy}
    \end{subfigure}     
    \begin{subfigure}[t]{0.49\linewidth}
        \includegraphics[width=1.0\columnwidth]{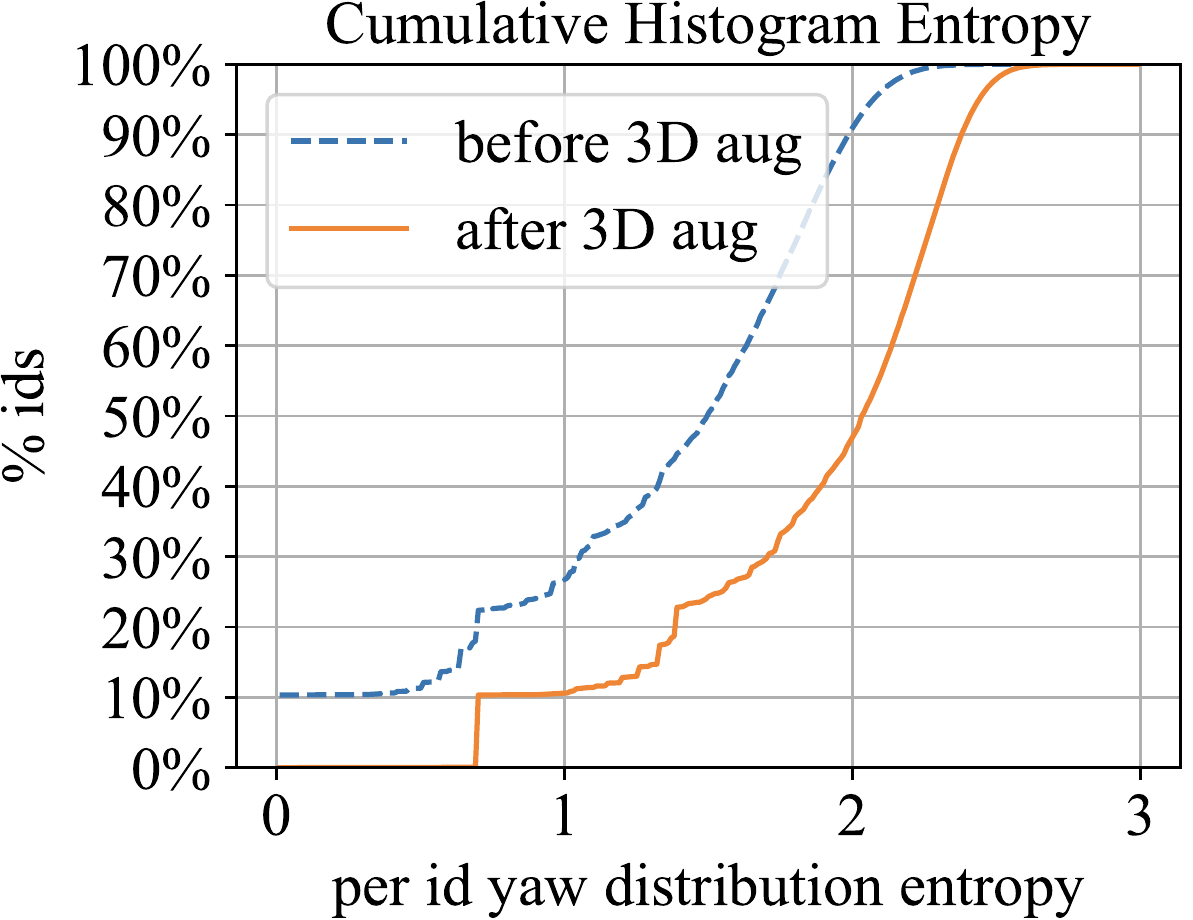} 
        \caption{}\label{fig:entropy_before_after}
    \end{subfigure}
    \caption{\label{fig:entropy_compare} (a): Absolute yaw angle distribution of a sample identity. (b): The cumulative entropy histogram (the more right, the better) over identities in TrillionPairs \cite{TrillionPairs} before and after applying 3D pose augmentation with the sampling method of yaw entropy cut-off. After 3D data augmentation, the yaw pose distribution is much richer.}
\end{figure}
% This identity has an entropy value of $2.10$

\paragraph{IJBC 1:1 Covariate Benchmark} is a protocol used for studying image pair verification. We use it to conduct in-depth breakdown analysis of models trained with and without the proposed 3D data synthesis pipeline. \tabref{tbl:compare_ijbc_covarite_sota} illustrates that models trained with the proposed 3D data augmentation method achieve the state-of-art and significantly improve over the baselines, such as a 30\% absolute TAR increase at FAR1e-4 when trained on TrillionPairs\cite{TrillionPairs}. In addition, on IMDB \cite{wang2018devil} training, significant performance boost of absolute $7\%$-$8\%$ is achieved at large yaw pose groups for low FAR regions. This demonstrates the superiority of the proposed method in enhancing the robustness of face recognition model training against pose variation.

\paragraph{IJBA 1:1 Verification Breakdown analysis} \figref{fig:ijba_plot_compare_yaw} shows the breakdown analysis of 1:1 template verification on IJBA against the yaw pose group of input face images. Models are trained on the IMDB \cite{wang2018devil} dataset. The maximum yaw angle of images in each template is used. And the template pair pose statistic is taken as the average of angle statistics of two templates. It shows that models trained with 3D pose augmentation is more robust than baselines against pose variation, especially at large viewpoints.

\begin{figure}[tbh]
    \begin{subfigure}[tbh]{0.49\linewidth}
        \includegraphics[width=1.0\columnwidth]{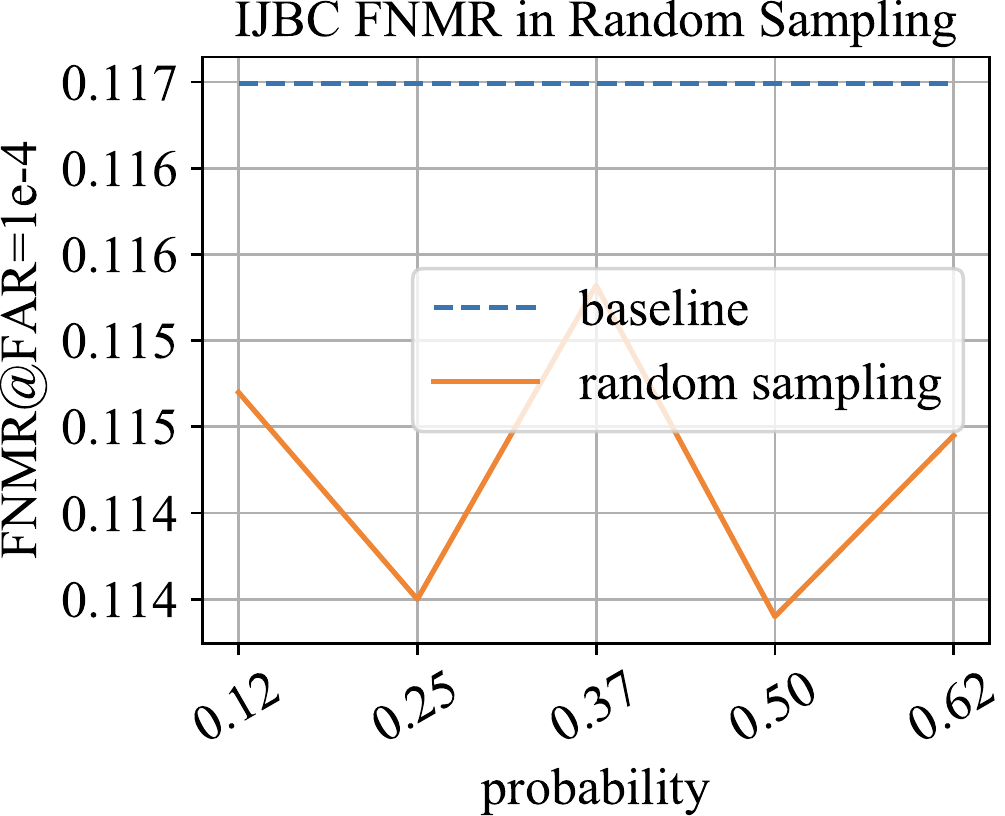} 
        \caption{}
    \end{subfigure}
    \begin{subfigure}[tbh]{0.49\linewidth}
            \includegraphics[width=1.0\columnwidth]{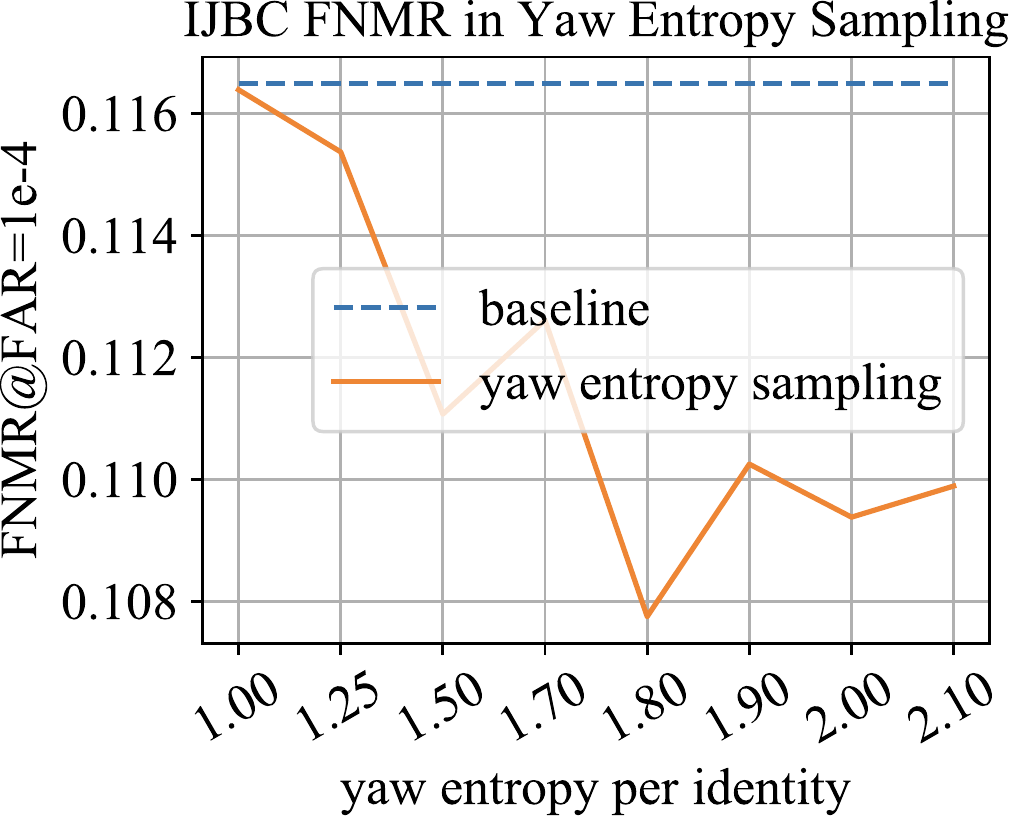}
            \caption{}
    \end{subfigure}
    \begin{subfigure}[tbh]{0.49\linewidth}
        \includegraphics[width=1.0\columnwidth]{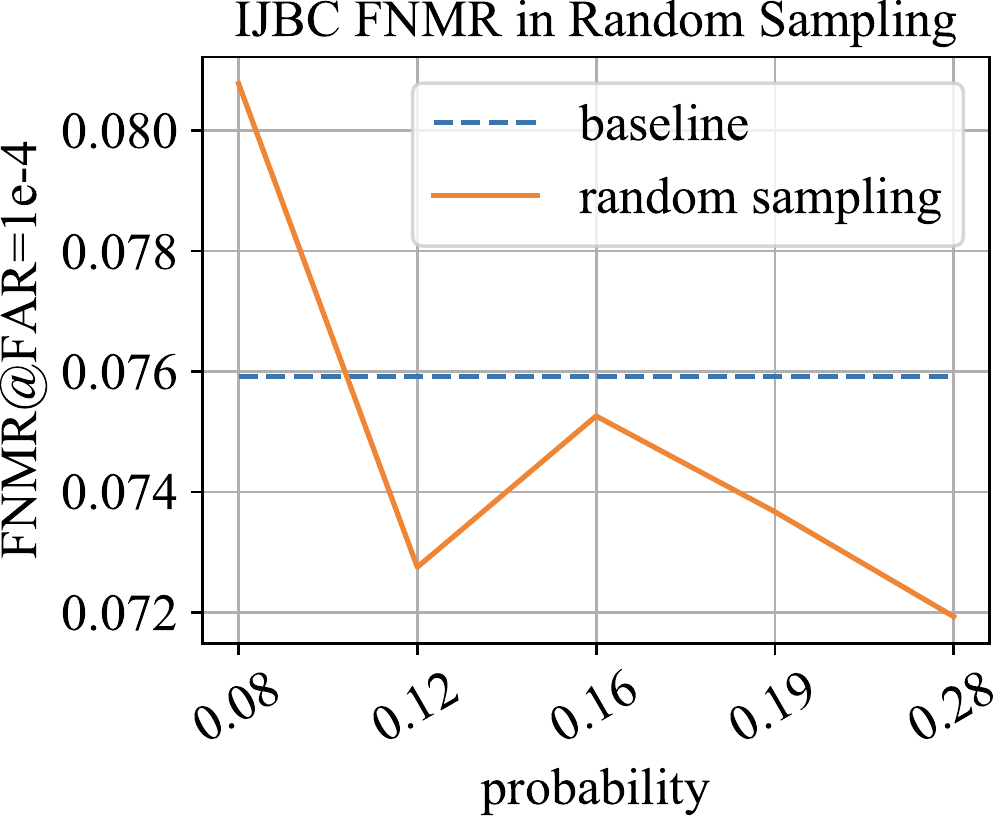} 
        \caption{}
    \end{subfigure}
    \begin{subfigure}[tbh]{0.49\linewidth}
        \includegraphics[width=1.0\columnwidth]{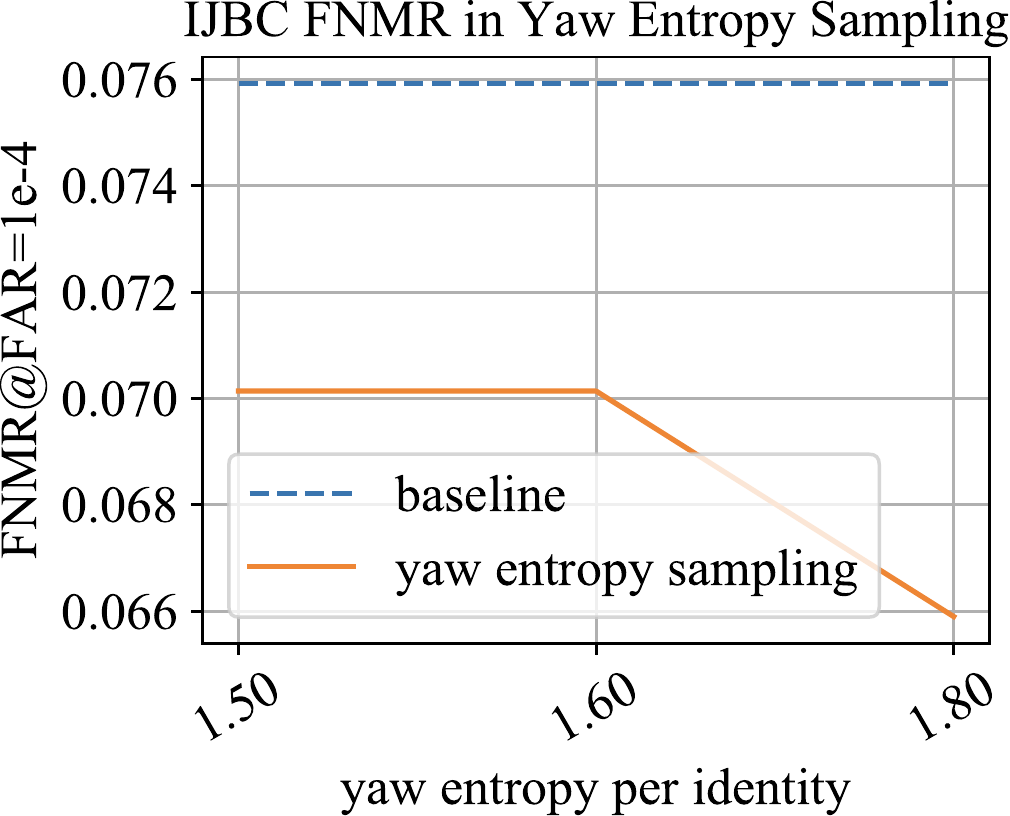}
        \caption{}
    \end{subfigure}
    \caption{\label{fig:plot_sample} IJBC \cite{ijbc} template Verification FNMR (lower better) at FAR = $1e^{-4}$ of models trained with and without 3D pose data augmentation on IMDB\cite{wang2018devil} (a,b) and TrillionPairs\cite{TrillionPairs} (c, d) using different sampling methods: 1) Random (a,c). 2) Per identity yaw pose entropy cut-off sampling (b,d). Both sampling methods bring significant error reduction to the baselines. The second method using the yaw pose entropy of identities achieves more significant error reduction than random sampling on both training sets.}
\end{figure}

\subsection{Ablation Studies\label{sec:ablation_experiment}}

\paragraph{Impact of Image Sampling in Augmentation} When applying the proposed 3D data augmentation method, we experimented with different augmentation strategies. As mentioned in \secref{sec:3D_to_face_analysis}, we designed two sampling methods for 3D-aided face recognition: 1) random sampling 2) a sampling method using the entropy of pose distribution of the training identities as defined in Equation~\ref{eq:yaw_entropy}. We evaluate these two image sampling methods with the IJBC template verification protocol and the metric of FNMR at FAR of $1e^{-4}$. \figref{fig:plot_sample} shows the comparison of models trained with and without pose 3D augmentation using these two methods on the IMDB \cite{wang2018devil} and TrillionPairs \cite{TrillionPairs} dataset. Though both sampling methods significantly reduce the FNMR from the baselines, the second method of identity yaw entropy cut-off sampling outperforms random sampling in all training scenarios. This validates the effectiveness of our hypothesis in using this sampling method. \figref{fig:entropy_compare} shows an exemplar identity's yaw distribution and the yaw entropy distribution over all identities in TrillionPairs before and after 3D pose augmentation. The variation of yaw pose distributions is much richer after 3D pose augmentation than before.

\begin{table}[tbh]
\begin{center}
\begin{footnotesize}
\setlength{\tabcolsep}{0pt}
\centering
\begin{minipage}[t]{1.0\linewidth}
        \begin{adjustbox}{max width=\linewidth}
    \begin{tabular}{>{\centering\arraybackslash}p{2.4cm}|>{\centering\arraybackslash}p{0.96cm}>{\centering\arraybackslash}p{0.96cm}|>{\centering\arraybackslash}p{1.31cm}|>{\centering\arraybackslash}p{0.93cm}>{\centering\arraybackslash}p{0.93cm}>{\centering\arraybackslash}p{0.93cm}>{\centering\arraybackslash}p{0.93cm}>{\centering\arraybackslash}p{0.93cm}>{\centering\arraybackslash}p{0.93cm}}
        Model Training & \multicolumn{2}{c}{1:1 Verif TAR@FAR} & \multicolumn{1}{c}{1:N TPIR} & \multicolumn{6}{c}{1:1 Covariate TAR@FAR1e-3 Yaw Group}\\[0.5ex] 
        \hline
                                             & 1e-3 & 1e-4 & FPIR0.1 & [0,10] & [10,30] & [30,50]& [50,70] & [70,90] & All \\
    	\hline\hline
    	Baseline & 0.939  &  0.883 & 0.885 & 0.959 &  0.937& 	0.883& 0.799& 0.677& 0.890\\
        \hline
        Aug 2D-similarity & 0.938 & 0.886 & 0.888 &  0.959 &	0.941	& 0.884&	0.781	&0.645& 0.891\\
                               \hline
        Aug 3D Pose (Ours)  & \textbf{0.945}  &  \textbf{0.892} & \textbf{0.895} &	\textbf{0.960} & \textbf{0.945} & \textbf{0.910}& \textbf{0.831} &\textbf{0.713} & \textbf{0.908} \\
                               \hline
    \end{tabular}
    \end{adjustbox}
    \end{minipage}
    % \vspace{-1em}
\end{footnotesize}
    \caption{\label{tbl:2D_simi} IJBC \cite{ijbc} template verification, identification and 1:1 Covariate verification comparison of our method with 2D similarity transform augmentation; Models are trained on IMDB \cite{wang2018devil}}
\end{center}
\end{table}

\begin{table}[tbh]
\begin{center}
\begin{footnotesize}
\setlength{\tabcolsep}{0pt}
\centering
\begin{adjustbox}{max width=\linewidth}
    \begin{tabular}{>{\centering\arraybackslash}p{2.6cm}>{\centering\arraybackslash}p{0.9cm}>{\centering\arraybackslash}p{1.0cm}>{\centering\arraybackslash}p{1.0cm}|>{\centering\arraybackslash}p{1.0cm}>{\centering\arraybackslash}p{1.0cm}>{\centering\arraybackslash}p{0.8cm}}
        Model Training & \multicolumn{3}{c}{1:1 Verif TAR@FAR}& \multicolumn{2}{c}{1:N TPIR@FPIR} & Rank \\
        \hline
    	 & 1e-2 & 1e-3 & 1e-4 & 0.1 & 0.01 & 1 \\ [0.5ex] 
    	\hline\hline
    	DR-GAN\cite{tran2017disentangled} & 0.774 & 0.539 & - & - & - & 0.855 \\
    	\hline
    	FF-GAN \cite{yin2017towards} & 0.852& 0.663 & - & - & 0.902 \\
    	\hline
    	LB-GAN \cite{LB-GAN}  & 0.923 & 0.804 & - & - & - & 0.947 \\
    	\hline
    	DA-GAN\textsubscript{2.0} \cite{zhao20183d} & \textbf{0.989} & \textbf{0.973} & 0.946 & \textbf{0.982} & 0.939 & \textbf{0.990} \\ 
    	\hline
    	\hline
    	Ours & 0.984 & \textbf{0.973} &	\textbf{0.959} & 0.976 & \textbf{0.947} & 0.985 \\
    	\hline
    \end{tabular}
\end{adjustbox}
        \caption{\label{tbl:compare_GAN} IJBA \cite{ijba} template verification and identification comparison of our method with other GAN based methods.}
\end{footnotesize}
\end{center}
\end{table}

\begin{table}[tbh]
\begin{center}
\begin{footnotesize}
\setlength{\tabcolsep}{1pt}
\centering
\begin{minipage}[t]{1.0\linewidth}
        \begin{adjustbox}{max width=\linewidth}
    \begin{tabular}{>{\centering\arraybackslash}p{2.7cm}|>{\centering\arraybackslash}p{1.0cm}>{\centering\arraybackslash}p{1.0cm}>{\centering\arraybackslash}p{1.0cm}|>{\centering\arraybackslash}p{1.3cm}|>{\centering\arraybackslash}p{1.0cm}>{\centering\arraybackslash}p{1.0cm}>{\centering\arraybackslash}p{1.0cm}>{\centering\arraybackslash}p{1.0cm}}
    	Model Training & \multicolumn{3}{c}{IJBC 1:1 Verif TAR@FAR}& 1:N TPIR & \multicolumn{4}{c}{IJBC Covariate 1:1 Verif TAR@FAR}\\[0.5ex] 
    	\hline
    	 & 1e-2 & 1e-3 & 1e-4 & FPIR0.1 & 1e-1 & 1e-2 & 1e-3 & 1e-4 \\ [0.5ex] 
    	 \hline\hline
    	Baseline & 0.973 & 0.939 & 0.883 & 0.885 & 0.984 & 0.953 & 0.890 & 0.728 \\ 
    	\hline
    	Generic 3D A.Pose & 0.974 & 0.944 & 0.890 & 0.892 & 0.986 & 0.959 & 0.900 & 0.739\\
    	\hline
    	Img-dep A.Pose & \textbf{0.976} & \textbf{0.945} & \textbf{0.892} & \textbf{0.895}& \textbf{0.987} &  \textbf{0.962} &  \textbf{0.908} & \textbf{0.774}\\
    	\hline
    \end{tabular}
    \end{adjustbox}
    \end{minipage}
        \caption{\label{tbl:compare_generic_ijba} IJBC \cite{ijbc} 1:1 template verification, identification and 1:1 Covariate Image verification comparison of data augmentation with an image-dependent 3D reconstruction method compared to with a generic 3D face; Models are trained on IMDB \cite{wang2018devil}}
\end{footnotesize}
\end{center}
\end{table}

\begin{table}[tbh]
\begin{center}
\begin{footnotesize}
\setlength{\tabcolsep}{0pt}
\centering
\begin{adjustbox}{max width=\linewidth}
    \begin{tabular}{>{\centering\arraybackslash}p{2.6cm}>{\centering\arraybackslash}p{0.9cm}>{\centering\arraybackslash}p{1.0cm}>{\centering\arraybackslash}p{1.0cm}|>{\centering\arraybackslash}p{1.0cm}>{\centering\arraybackslash}p{1.0cm}>{\centering\arraybackslash}p{0.8cm}}
        Model Training & \multicolumn{3}{c}{1:1 Verif TAR@FAR}& \multicolumn{2}{c}{1:N TPIR@FPIR} & Rank \\
        \hline
    	 & 1e-2 & 1e-3 & 1e-4 & 0.1 & 0.01 & 1 \\ [0.5ex] 
    	\hline\hline
    	Masi et al.\cite{MasiEtAl:MillionsFaces:ECCV:2016} & 0.866 & 0.636  & - & - & - & 0.872 \\   
    	 \hline
        Lv et al. \cite{dataset-augmentation} & - &0.936 & 0.790 & 0.740 & 0.586 & 0.840\\
    	\hline
    	Crispell et al.\cite{crispell2017dataset} & - & - & - & 0.870 & 0.734 & 0.944 \\
        \hline
    	\hline
    	Ours & 0.984 & \textbf{0.973} &	\textbf{0.959} & \textbf{0.976} & \textbf{0.947} & \textbf{0.985} \\
    	\hline
    \end{tabular}
\end{adjustbox}
        \caption{\label{tbl:compare_3D} IJBA \cite{ijba} template verification and identification results comparison of our method with other 3D model based augmentation methods.}
\end{footnotesize}
\end{center}
\end{table}

% \subsubsection{Impact of face alignment in the face recognition pipeline.}
% As shown above the proposed 3D face augmentation pipeline can be used to improve face landmark localization, especially at large yaw and pitch view points, over the metric of normalized mean error. This enable us to perform a study on the  impact of face landmark localization quality to the final face recognition performance. 
% We experiment the combination of whether we use the data augmentation in training both the landmark localization model and the embedding models. The results are show in \tabref{tbl:compare_ijba}. We can observe that with a more robust face alignment as pre-processing, the performance of facial recognition system improve significantly. 
% Intestingly, when augmentation is enabled on both components, the accuracy were not further improved. This could be partly due to the augmentation artifacts accumulated over the pipeline.  We leave the investigation of this observation for future works. 

\paragraph{Compare with 2D Based Data Augmentation}
\tabref{tbl:2D_simi} shows the performance of a model trained on IMDB \cite{wang2018devil} with additional 2D similarity transform augmentation including 2D image rotation and translation besides the basic ones used in the baseline for face recognition on IJBC \cite{ijbc} benchmark. It shows that such 2D similarity transform augmentation is not as effective in improving model robustness compared to ours.

\paragraph{Compare with GAN Based Methods}
\tabref{tbl:compare_GAN} shows the performance comparison over IJBA \cite{ijba} benchmark of the proposed 3D data augmentation method with other GAN based methods with representation learning by synthesis or data augmentation for face recognition. It shows despite the simplicity of our method which does not involve adversarial training, we out-perform other GAN based approaches.

\paragraph{Compare with Generic 3D Based Augmentation \label{sec:compare_3d_aug}}
To study the impact of different 3D face modeling methods in our framework to the final face understanding performance. We perform an ablation study in swapping the image-dependent volumetric 3D face reconstruction method used in the previous experiments into a generic parametric face shape defined as the mean face in \cite{bfm09}. \tabref{tbl:compare_generic_ijba} gives the results of this ablation study on face recognition. It shows that though our framework is also able to get better accuracy and robustness than the baselines with a generic 3D shape, an image-dependent shape reconstruction gives more performance boost, especially for the challenging 1:1 IJBC Covaraite benchmark. Visual examples of augmented images using these two different 3D modeling techniques together with their perceptual quality (inception-score) are included in section \ref{sec:compare_vis_3D} of the appendix. We also compare our approach with other 3D data augmentation methods for face recognition with a set of generic 3D faces \cite{MasiEtAl:MillionsFaces:ECCV:2016} or sparsely fitted 3D models \cite{crispell2017dataset, dataset-augmentation} in \tabref{tbl:compare_3D}. It is observed that our framework achieves superior results than these methods.

\section{Conclusions\label{sec:conclusions}}

We described a method to generate realistic 3D augmented face images that increases the robustness of a given face understanding system on the tasks of face landmark localization, face attributes classification and face recognition. It utilizes an underlying 3D modeling procedure to increase the 3D rigid pose and illumination variation of training datasets to improve the robustness of trained deep neural networks. 

% Future work includes additional constraints during face-analysis model training. From the breakdown analysis of models trained with 3D pose data augmentation against the rotation angle of input face images, while there is significant performance boost for images displaying large viewpoints, occasionally, there can be some small degradation of performance at near-frontal views, for instance, for gender classification. To overcome this, techniques of knowledge distillation and learning without forgetting \cite{hinton2015distilling, li2017learning} could be applied during the training stages for these face analysis tasks.

%%% Local Variables:
%%% mode: latex
%%% TeX-master: "data_augmentation_for_face_recognition"
%%% End:

\clearpage
\appendix
\section{Appendix}
\subsection{Details of Texture Mapping\label{sec:texture_map_appendix}}
We use vertex-coloring and transfer the original image's RGB pixel values onto the 3D mesh vertices. We assume an orthographic camera projection and align the 3D mesh to the camera-centered coordinate system of the input image using the 6dof pose of the 3D model estimated. Texture mapping is done by projecting each mesh vertex onto the image plane and assigning the RGB of its nearest pixel, namely, mapping a vertex (x, y, z) to image coordinates (x, y), and finding pixel (u, v) on the original input image closest to (x, y).

\subsection{Details of Rendering Setup \label{sec:rendering_setup_appendix}}
\begin{figure}[tbh]
    \begin{subfigure}[tbh]{0.25\linewidth}
        \includegraphics[width=1.\columnwidth]{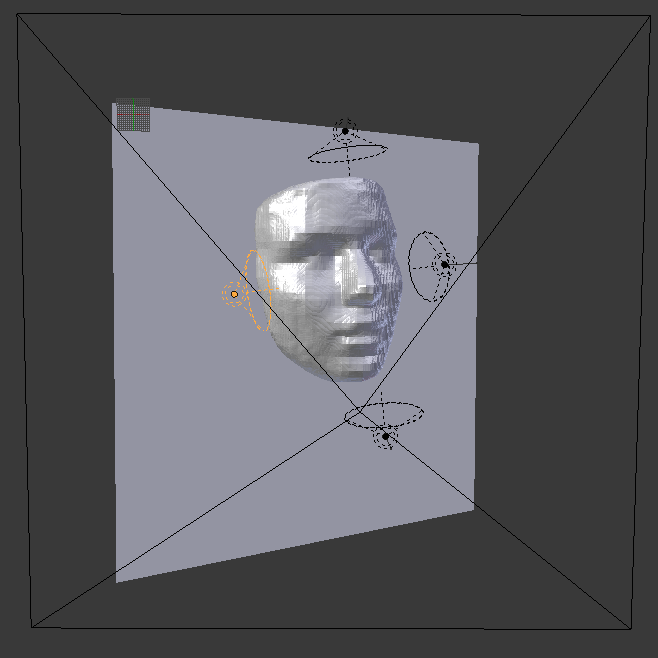}
        \caption{}\label{fig:blender_setup}
    \end{subfigure}
    \begin{subfigure}[tbh]{0.74\linewidth}
        \includegraphics[width=0.24\columnwidth]{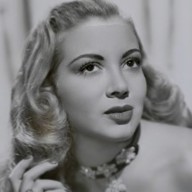} 
        \includegraphics[width=0.24\columnwidth]{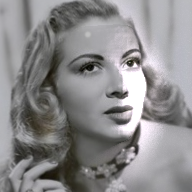}
        \includegraphics[width=0.24\columnwidth]{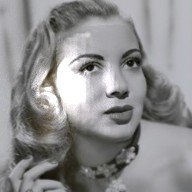}
        \includegraphics[width=0.24\columnwidth]{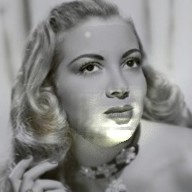}
        \caption{}\label{fig:example_render}
    \end{subfigure}
    \caption{\label{fig:render_setup} Illumination augmentation Blender rendering setup. Four additional light sources are shown in (a): to the top, left, right and bottom of the 3D face shape estimated. (b) original input image (left-most) and exemplar relighted images, each with one of these light sources toggled on.}
\end{figure}

Here we provide details of the rendering setup that produces  ~\emph{large volume} and ~\emph{high variance} 3D pose and illumination augmented images. We use Blender \cite{Blender} for this rendering process. As shown in \figref{fig:blender_setup}, given the estimated 3D face shape and the background plane, we place four additional light sources of type ~\emph{Spot} around the top, left, right and bottom side of the 3D face shape. These light sources are placed in the world coordinate system aligned with the 3D face shape. When combining the pose rotation and illumination augmentation, we apply the same rigid-body transformation to these four light sources as the 3D face shape. In rendering the illumination augmented images, one of these four additional light sources will be randomly toggled on. For this process, we adopt a mix shader combining an emissive illumination model and the diffuse model utilizing the bidirectional scattering distribution function \cite{asmail1991bidirectional} (BSDF) to estimate the probability of ray reflectance at the surface of the 3D face mesh model. Some sample relighted images using one of these light sources are shown in \figref{fig:example_render}.

\subsection{Details of Pose Augmentation Constraints\label{sec:rigid_control_appendix}}
Here we provide the details of constraints on the rigid-rotation of the 3D face model to produce high-quality realistic 3D pose augmented images. We constrain the rotation angles around y (yaw) not to expose self-occluded regions of the 3D mesh. This is achieved by rotating the model away from the bilateral symmetry plane of the 3D face model (noted with $[\normal_{\text{bilateral}}^{T}, d_{\text{bilateral}}]$). In addition, we constrain the rotation around y (yaw) and x (pitch) up to the maximal angle such that the face gaze direction does not exceed $90^{\circ}$ w.r.t the camera viewing direction. Specifically, the norms of the back plane ($\normal_{\text{backplane}}$) and the bilateral symmetry plane ($\normal_{\text{bilateral}}$) of the 3D face mesh are checked against the viewing direction ($(0, 0, 1)$ in the camera coordinate system of the original input image, assuming an orthographic camera model) using the following conditions to avoid exposing the back of the face mesh and the self-occluded region. 
% {\color{red} give a plot demo}
\begin{eqnarray}
	[Q \normal_{\text{bilateral}} \times (0, 0, 1)]_{y} \geqslant 0 \nonumber\\
	Q \normal_{\text{backplane}} \cdot (0, 0, 1) \leqslant 0 
\end{eqnarray} 
Where $Q = \begin{psmallmatrix}\mathbf{R} & \mathbf{t}\\ \mathbf{0^{T}} & 1\end{psmallmatrix}$ is the rigid body transformation for the pose change. Here, $\mathbf{R}$ is the combination of the pitch and yaw rotation, and $[\cdot]_{y}$ is the y component of the vector in $\mathbb{R}^{3}$

\begin{table}[tbh]
\begin{center}
\begin{footnotesize}
\setlength{\tabcolsep}{0pt}
\centering
\begin{adjustbox}{max width=\linewidth}
    \begin{tabular}{>{\centering\arraybackslash}p{3.9cm}>{\centering\arraybackslash}p{0.9cm}>{\centering\arraybackslash}p{1.0cm}>{\centering\arraybackslash}p{1.0cm}|>{\centering\arraybackslash}p{1.0cm}>{\centering\arraybackslash}p{1.0cm}>{\centering\arraybackslash}p{0.8cm}}
        Model Training & \multicolumn{3}{c}{1:1 Verif TAR@FAR}& \multicolumn{2}{c}{1:N TPIR@FPIR} & Rank \\
        \hline
    	 & 1e-2 & 1e-3 & 1e-4 & 0.1 & 0.01 & 1 \\ [0.5ex] 
    	\hline\hline
    	B. IMDB & 0.983 & 0.955 & 0.821 & 0.972 & 0.903 & 0.987\\ 
    	\hline
    	 A.Illum IMDB & 0.985 & 0.958 & 0.833 & 0.974 & 0.906 & 0.988 \\
    	\hline
    	A.Pose IMDB & \textbf{0.986} & 0.961 & 0.851 &0.974 & 0.913 & \textbf{0.990} \\
    	\hline
    	A.Pose+A.Illum IMDB & \textbf{0.986} & 0.959 & 0.838 & 0.974 & 0.906 & 0.989  \\
    	\hline
    	\hline
        B. TrillionPairs &  0.983 &  0.969 & 0.948 & 0.974 &  0.945 & 0.982 \\
        \hline
        A.Illum TrillionPairs & 0.981 & 0.968 & 0.944 & 0.972 & 0.942 & 0.980\\
        \hline
        A.Pose TrillionPairs & 0.984 & 0.971 & 0.958 & 0.975 & 0.947 & 0.982 \\
        \hline
        A.Pose+A.Illu TrillionPairs & 0.983& 0.971 & 0.947 & 0.974 & 0.944 & 0.980\\
        \hline
        \hline
        B. TrillionPairs*& 0.984 & 0.972 & \textbf{0.959}  & 0.975	& 0.939 & 0.984 \\
    	\hline
    	A.Illum TrillionPairs*& 0.982 & 0.971 & 0.956 & 0.973 & 0.936 & 0.981 \\
    	\hline
    	A.Pose TrillionPairs* & 0.984 & \textbf{0.973} &	\textbf{0.959} & \textbf{0.976} & \textbf{0.947} & 0.985 \\
    	\hline
    	A.Pose+A.Illu TrillionPairs*&0.983 & 0.972 & 0.954 & 0.973 & 0.939& 0.984 \\
    	\hline
    \end{tabular}
\end{adjustbox}
        \caption{\label{tbl:compare_ijba_ablation} Comparison of models trained on different datasets and different 3D augmentation on IJBA \cite{ijba} Training results on IMDB \cite{wang2018devil} and TrillionPairs \cite{TrillionPairs} dataset are shown. B. IMDB and B. TrillionPairs are baseline models trained on IMDB and TrillionPairs, respectively. A.Illum and A.Pose refer to illumination and pose 3D augmented model training. * refers to Media-Pooling as in \cite{DBLP:journals/corr/abs-1809-07586}}
\end{footnotesize}
\end{center}
\end{table}

\subsection{Full Results on IJBA and IJBC \label{sec:full_results}}
Here, in \tabref{tbl:compare_ijba_ablation} and \tabref{tbl:compare_ijbc}, we provide the full results of face template verification and identification with training using different datasets and different 3D augmentation over the IJBA \cite{ijba} and IJBC \cite{ijbc} benchmark. We further provide in \tabref{tbl:compare_ijbc} the IJBC \cite{ijbc} results at low FAR and FNIR thresholds. It is observed that with the proposed 3D data augmentation, substantial performance improvements over baselines are achieved across all FAR and FNIR thresholds. On the IJBC benchmark, at low FAR and FNIR thresholds where high prediction precision is required, we achieve significant performance increase of an absolute 4-6\% in face verification and an absolute 15-20\% in face identification when training on TrillionPairs \cite{TrillionPairs} dataset.

\begin{table}[h]
\begin{center}
\begin{footnotesize}
\setlength{\tabcolsep}{0pt}
\centering
\begin{adjustbox}{max width=\linewidth}
    \begin{tabular}{>{\centering\arraybackslash}p{3.0cm}>{\centering\arraybackslash}p{0.9cm}>{\centering\arraybackslash}p{1.0cm}>{\centering\arraybackslash}p{1.0cm}>{\centering\arraybackslash}p{1.0cm}|>{\centering\arraybackslash}p{1.0cm}>{\centering\arraybackslash}p{1.0cm}>{\centering\arraybackslash}p{1.0cm}>{\centering\arraybackslash}p{1.0cm}>{\centering\arraybackslash}p{0.8cm}}
        Model Training & \multicolumn{4}{c}{1:1 Verif TAR@FAR}& \multicolumn{4}{c}{1:N TPIR@FPIR} & Rank \\
        \hline
        & 1e-2 & 1e-3 & 1e-4 & 1e-5 & 0.1 & 0.01 &0.001 & 0.0001 & 1 \\ [0.5ex] 
    	\hline\hline
    	B. IMDB & 0.973 &	0.939	&0.883 & 0.809 & 0.885&	0.799 &0.672 & 0.509 & 0.929 \\ 
    	\hline
    	A.Illum IMDB  &0.974	&0.940&	0.886 & 0.811 &0.888&	0.805 & 0.697	& 0.549& 0.931 \\
    	\hline
    	A.Pose IMDB &0.976&	0.945&	0.892 & 0.813& 0.895	&0.807 & 0.689 &	0.525& 0.936 \\
    	\hline
    	A.Pose + A.Illum IMDB &0.977 & 0.943&0.888 &0.814 & 0.892 & 0.811 & 0.708 & 0.530 & 0.933  \\
    	\hline
    	\hline
        B. TrillionPairs  & 0.977 &	0.958	& 0.921 & 0.827 &0.924	& 0.829 & 0.621 &	0.467 & 0.933\\
        \hline
        A.Illum TrillionPairs  & 0.976	&0.956 &0.922& 0.853 &0.923	& 0.867 &0.706 &0.603 & 0.937\\
        \hline
        A.Pose TrillionPairs  & 0.979	&0.962	&0.934 & 0.889 &0.933	& 0.888 & 0.781 &	0.688 & 0.945\\
        \hline
        A.Pose + A.Illum TrillionPairs  & 0.977 &0.959 &0.925 & 0.845& 0.926 & 0.843 &0.658 & 0.497 & 0.938\\
        \hline
        \hline
        B. TrillionPairs* & 0.983 &	0.970 &	0.947 & 0.888 &	0.949	 & 0.902 & 0.736&	0.553 & 0.956 \\
    	\hline
    	A.Illum TrillionPairs* &0.982 &0.968 &0.945 &0.906 & 0.945 & 0.910& 0.795& 0.684& 0.957\\
    	\hline
    	A.Pose TrillionPairs* &\textbf{0.984}& \textbf{0.972}	& \textbf{0.953} & \textbf{0.925}	 & \textbf{0.952}& \textbf{0.925} & \textbf{0.846} &	\textbf{0.749} &\textbf{0.963}\\
        \hline
    	A.Pose + A.Illum TrillionPairs* & 0.983& 0.970&0.949 & 0.904 & 0.949 & 0.904& 0.751& 0.583& 0.957\\
    	\hline
    \end{tabular}
\end{adjustbox}
\end{footnotesize}
    \caption{\label{tbl:compare_ijbc} Full performance results with low FAR and FNIR thresholds on IJBC \cite{ijbc} benchmark of models trained on IMDB \cite{wang2018devil} and TrillionPairs \cite{TrillionPairs} dataset. B. IMDB and B. TrillionPairs are baseline models trained on IMDB and TrillionPairs, respectively. A.Illum and A.Pose refer to illumination and pose 3D augmented model training, respectively. * refers to Media-Pooling as in \cite{DBLP:journals/corr/abs-1809-07586}}
\end{center}
\end{table}

% \subsection{Computation Complexity Compared to GAN}
% Here, we provide the details of computation complexity comparison of the proposed reconstruction and rendering paradigm based 3D data augmentation method as compared to a 2D GAN based method of DR-GAN for out-of-plane pose data augmentation. \tabref{tbl:runtime} shows the runtime and GPU memory consumption performance for DR-GAN and our 3D data augmentation running with a single NVIDIA TESLA V100 GPU. It is observed that while our method is able to achieve comparable runtime performance compared to DR-GAN, we require an order less memory footprint, demonstrating the superiority of our method in computational complexity.

% \begin{table}[bth]
% \begin{footnotesize}
% \begin{adjustbox}{max width=\linewidth}
% \setlength{\tabcolsep}{0pt}
% \centering
%     \begin{tabular}{|>{\centering\arraybackslash}p{2.4cm}|>{\centering\arraybackslash}p{4.2cm}|>{\centering\arraybackslash}p{2.9cm}|}
%         \hline
%     	Method & Runtime & GPU Memory Footprint \\ [0.5ex] 
%     	\hline
%     	DR-GAN & 14.6ms & 1124MiB\\ [0.5ex] 
%     	\hline
%     	Ours & 34.5ms & 225MiB \\ [0.5ex] 
%     	\hline
%     \end{tabular}
% \end{adjustbox}
% \caption{\label{tbl:runtime} Run-time and GPU memory footprint consumption comparison of our method to DR-GAN}
% \end{footnotesize}
% \end{table}

\subsection{Visual Comparison of Data Augmentation}
\begin{figure}[tbh]
\begin{adjustbox}{max width=\linewidth}
	\tabcolsep=0pt
\begin{tabu} to \textwidth {*7{X[1,c,m]}}
	original & $+10^\circ$ & $+20^\circ$ & $+40^\circ$ & $+10^\circ$ & $+20^\circ$ & $+40^\circ$\\ 
	\includegraphics[width=0.125\textwidth]{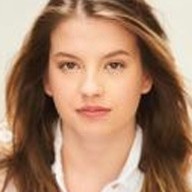}&
	\includegraphics[width=0.125\textwidth]{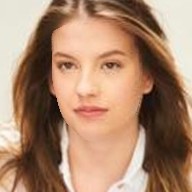}&
	\includegraphics[width=0.125\textwidth]{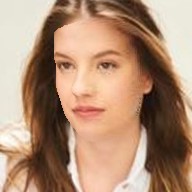}&
	\includegraphics[width=0.125\textwidth]{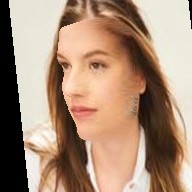}&
	\includegraphics[width=0.125\textwidth]{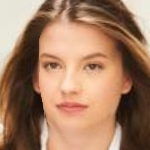}&
	\includegraphics[width=0.125\textwidth]{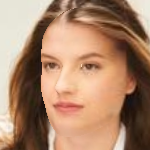}&
	\includegraphics[width=0.125\textwidth]{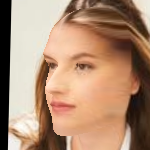}\\
	\includegraphics[width=0.125\textwidth]{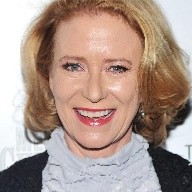}&
	\includegraphics[width=0.125\textwidth]{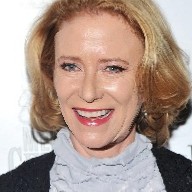}&
	\includegraphics[width=0.125\textwidth]{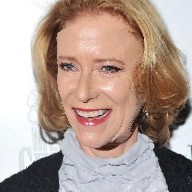}&
	\includegraphics[width=0.125\textwidth]{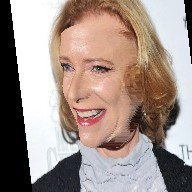}&
	\includegraphics[width=0.125\textwidth]{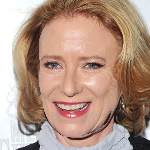}&
	\includegraphics[width=0.125\textwidth]{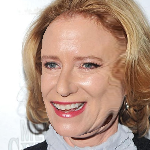}&
	\includegraphics[width=0.125\textwidth]{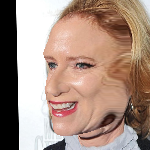}\\
	& \multicolumn{3}{c}{(a) VRN} & \multicolumn{3}{c}{ (b) Generic 3D} \\
\end{tabu}
\end{adjustbox}
\caption{\label{fig:synthesis_compare} 3D pose augmentation with (a) VRN ; (b) a generic 3D parametric model }
\end{figure}

\paragraph{Comparison of Different 3D Data Modeling Methods \label{sec:compare_vis_3D}}
Here, we show the visual comparison of 3D augmented images with different 3D modeling methods as mentioned in \secref{sec:compare_3d_aug} of the main paper. \figref{fig:synthesis_compare} shows some sample 3D pose augmented images on the IMDB \cite{wang2018devil} dataset. \tabref{tbl:synthesis_quality_quantitative} shows the Inception Scores (the higher, the better) of the complete set of 3D pose augmented images on IMDB using the two 3D modeling methods. It shows that the image-specific reconstruction of VRN \cite{JacksonEtAl:FaceRecon:ICCV:2017} produces more visually-appealing results than a generic 3D face mesh model \cite{bfm09}, which is also validated via the end-to-end face recognition results in \tabref{tbl:compare_generic_ijba} of the main paper.

\begin{figure}[tbh]
\begin{adjustbox}{max width=\linewidth}
    \includegraphics[width=1.0\textwidth]{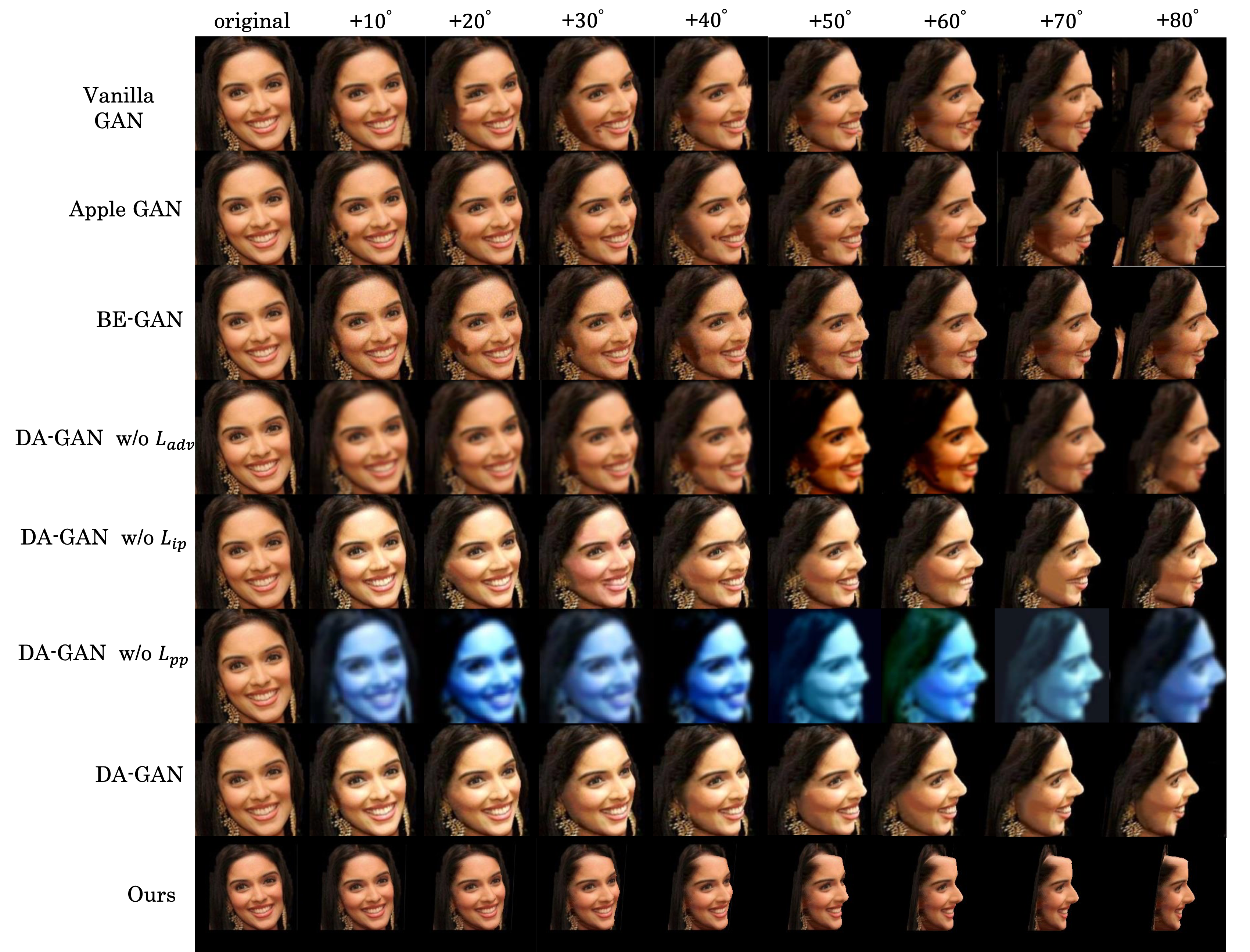}
\end{adjustbox}
    \caption{\label{fig:compare_gan} Our 3D pose augmentation compared to GAN based data augmentation methods. Our results are directly run on the image cropped from the paper \cite{zhao20183d}. Visual results of other GAN approaches are taken from \cite{zhao20183d}.}
\end{figure}

\paragraph{Comparison to other GAN based Data Augmentation} In \figref{fig:compare_gan}, we show the visual comparison of our 3D pose rotation augmented images with other 2D GAN based methods of Vanilla GAN, Apple GAN, BE-GAN and DA-GAN \cite{zhao20183d}. It is observed that despite the simplicity of our method which does not involve adversarial training, we get visually appealing results for pose rotation augmentation. For this visualization, we run our 3D pose data augmentation on the cropped image from \cite{zhao20183d}.

\begin{table}[bth]
\begin{footnotesize}
\begin{adjustbox}{max width=\linewidth}
\setlength{\tabcolsep}{0pt}
\centering
    \begin{tabular}{|>{\centering\arraybackslash}p{0.5cm}|>{\centering\arraybackslash}p{2.4cm}|>{\centering\arraybackslash}p{4.2cm}|>{\centering\arraybackslash}p{2.9cm}|}
% % no cite
% \begin{tabular}{|>{\centering\arraybackslash}p{0.4cm}|>{\centering\arraybackslash}p{2.1cm}|>{\centering\arraybackslash}p{3.5cm}|>{\centering\arraybackslash}p{2.1cm}|}
        \hline
    	IS & with VRN: 3.86 & with generic 3D model: 3.75 & StackGAN \cite{zhang2017stackgan}: 3.70 \\ [0.5ex] 
    	\hline
    \end{tabular}
\end{adjustbox}
\caption{\label{tbl:synthesis_quality_quantitative} Inception score of synthesized images on IMDB}
\end{footnotesize}
\end{table}

\subsection{Visualization of Face Recognition Results}
\begin{figure}[tbh]
\begin{adjustbox}{max width=\linewidth}
    \includegraphics[width=1.0\textwidth]{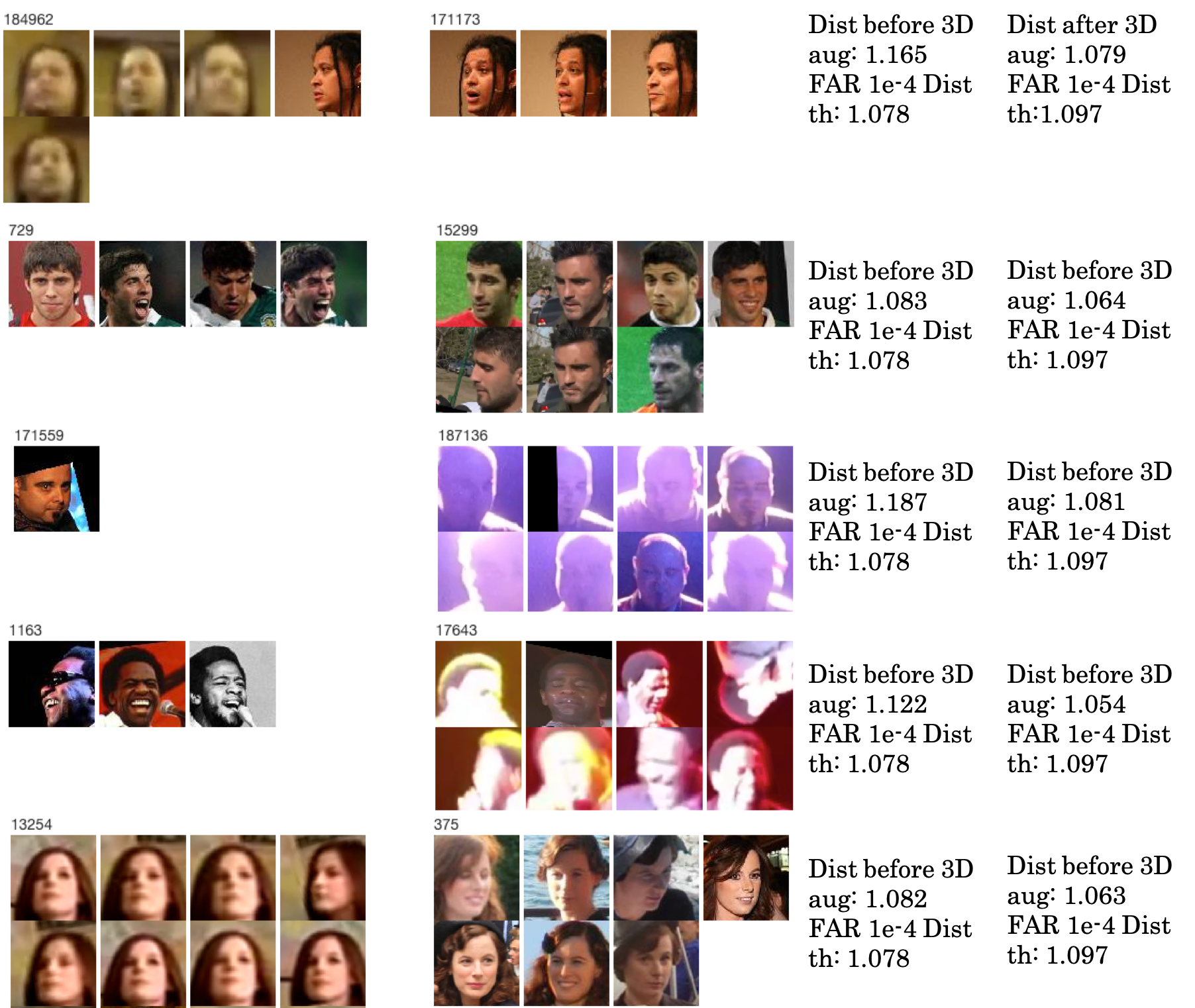}
\end{adjustbox}
    \caption{\label{fig:ijbc_compare_ref_positive} False non-match errors corrected after 3D pose augmentation on IJBC. Column 1 and 2 show the verification pairs and column 3 and 4 show the feature $l_{2}$ distance (Dist). Pairs shown are genuine examples of the same identity. Model specific $l_{2}$ distance threshold (Dist th) at FAR@1e-4 is shown, pairs having $l_{2}$ distance larger than it will be classified as imposter pair and vice versa.}
\end{figure}

\begin{figure}[tbh]
\begin{adjustbox}{max width=\linewidth}
    \includegraphics[width=1.0\textwidth]{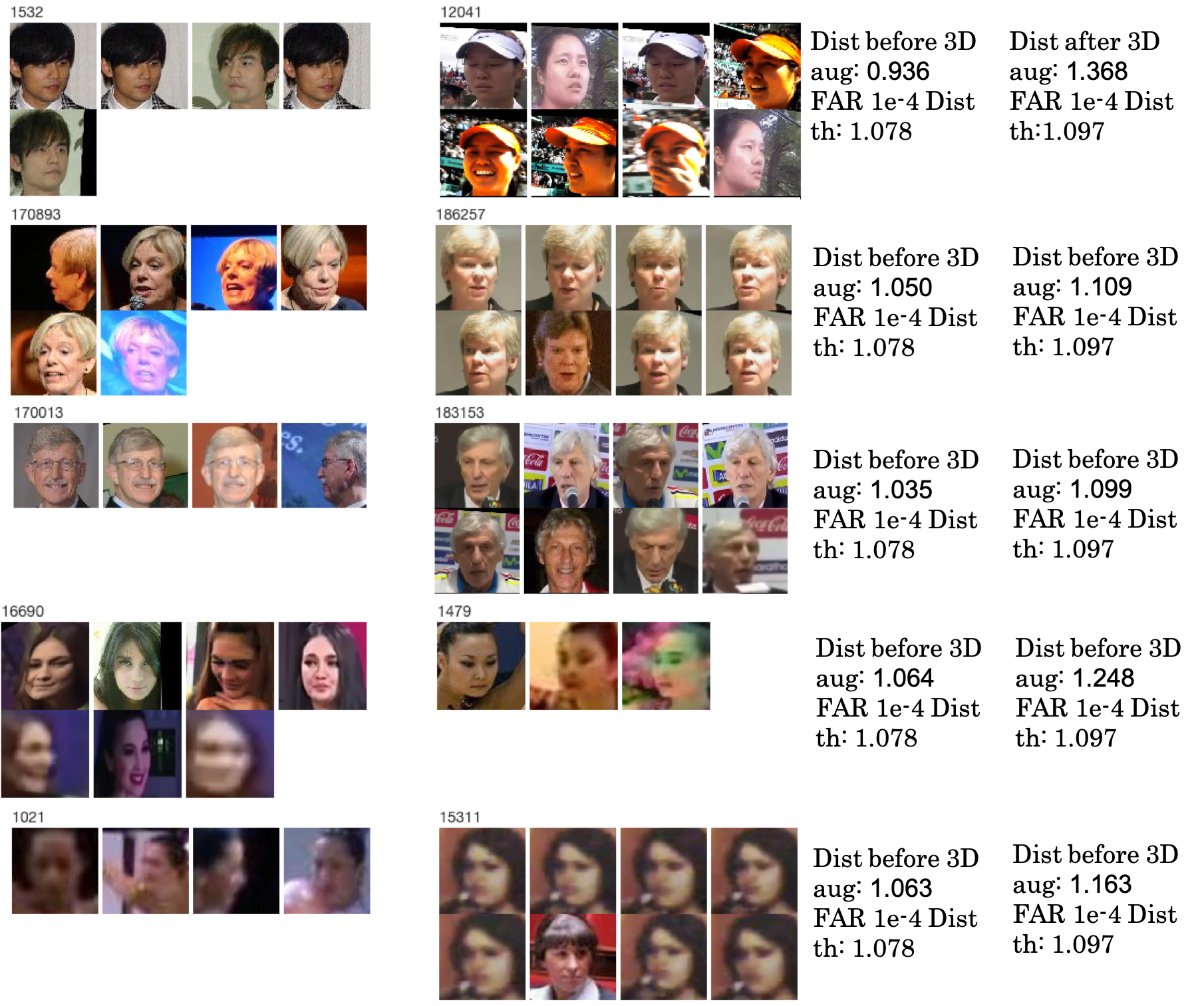}
\end{adjustbox}
    \caption{\label{fig:ijbc_compare_ref_negative} False match errors corrected after training the model with 3D pose augmentation on IJBC. Pairs shown are imposters of different identities. Model specific $l_{2}$ distance threshold (Dist th) at FAR@1e-4 is shown, pairs having $l_{2}$ distance smaller than it will be classified as genuine pair and vice versa. Errors made by model without 3D data augmentation are easily corrected with 3D pose augmentation.}
\end{figure}

Here, we provide the visualization of the template pairs used in the face verification task. We show the template pair verification feature $l_{2}$ distance from models trained with and without the proposed 3D data augmentation. A higher $l_{2}$ distance means the two template pairs are more dis-similar. We also show the threshold distance used for the evaluation metric of TAR@FAR1e-4 for the respective models. \figref{fig:ijbc_compare_ref_positive} and \figref{fig:ijbc_compare_ref_negative} shows such visualization from models trained with and without 3D pose data augmentation on the TrillionPairs \cite{TrillionPairs} dataset. \figref{fig:ijbc_compare_ref_positive} shows the genuine pairs (template pairs with the same identity, $l_{2}$ distance should be low) and \figref{fig:ijbc_compare_ref_negative} shows the imposter pairs (template pairs with different identity, $l_{2}$ distance should be high). It is observed that with the proposed 3D data augmentation, template pairs containing large pose variation are now more easily verified correctly with a much lower $l_{2}$ distance for genuine pairs and much higher $l_{2}$ distance for imposter pairs.

\subsection{Visualization of Face Landmark Localization}
\begin{figure}[tbh]
	\tabcolsep=0pt
\begin{center}
\begin{adjustbox}{max width=\linewidth}
\begin{tabu} to \textwidth {*4{X[1,c,m]}}
 0-5 yaw & 5-15 yaw & 15-40 yaw & $\geq$ 40 yaw \\ 
\includegraphics[width=0.24\textwidth]{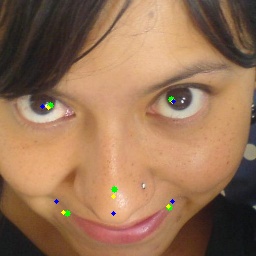}&
\includegraphics[width=0.24\textwidth]{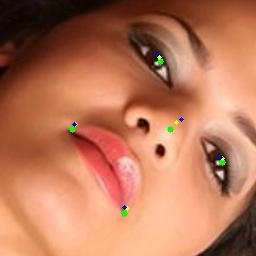}&
\includegraphics[width=0.24\textwidth]{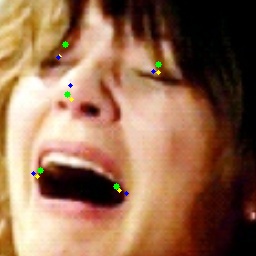}&
\includegraphics[width=0.24\textwidth]{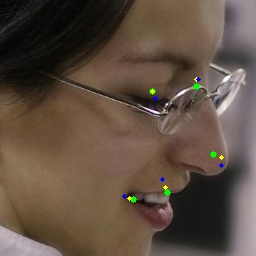}\\
\includegraphics[width=0.24\textwidth]{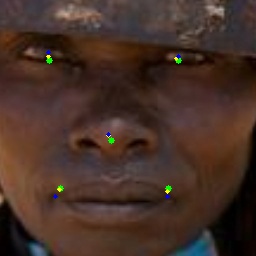}&
\includegraphics[width=0.24\textwidth]{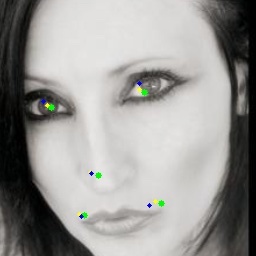}&
\includegraphics[width=0.24\textwidth]{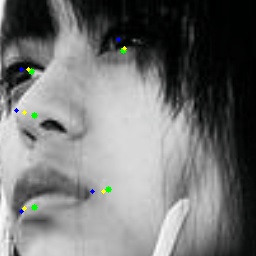}&
\includegraphics[width=0.24\textwidth]{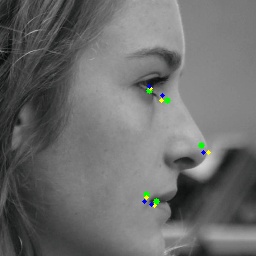}\\
\includegraphics[width=0.24\textwidth]{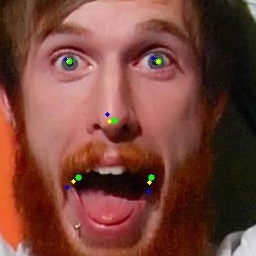}&
\includegraphics[width=0.24\textwidth]{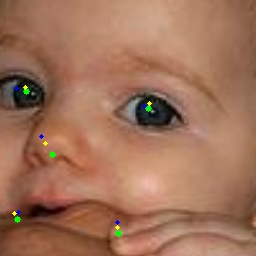}&
\includegraphics[width=0.24\textwidth]{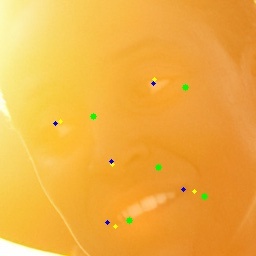}&
\includegraphics[width=0.24\textwidth]{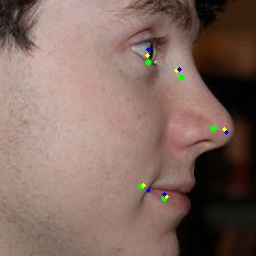}\\
\end{tabu}
\end{adjustbox}
\caption{\label{tbl:landmarks_yaw} Landmark visualization on images with different yaw angles. {\color{blue}Blue}: groundtruth; {\color{green}Green}: baseline; {\color{yellow}Yellow}: with 3D pose augmentation (Best viewed in color).}
\end{center}
\end{figure}

\begin{figure}[tbh]
\begin{center}
\begin{adjustbox}{max width=\linewidth}
	\begin{tabu} to \textwidth {*4{X[1,c,m]}}
	 $\geq$ 40 yaw & $\geq$ 40 yaw & $\geq$ 40 yaw & $\geq$ 40 yaw \\ 
\includegraphics[width=0.24\textwidth]{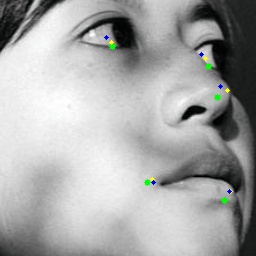}&
\includegraphics[width=0.24\textwidth]{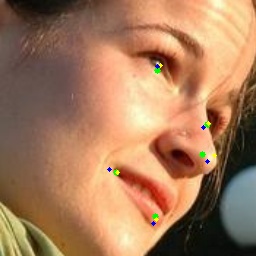}&
\includegraphics[width=0.24\textwidth]{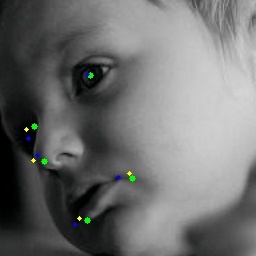}&
\includegraphics[width=0.24\textwidth]{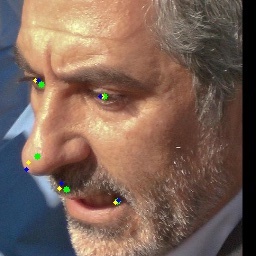}\\
\includegraphics[width=0.24\textwidth]{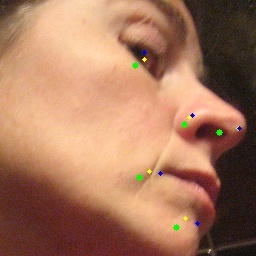}&
\includegraphics[width=0.24\textwidth]{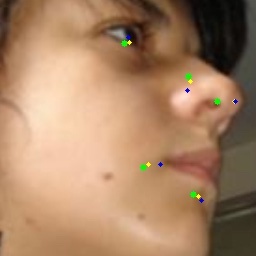}&
\includegraphics[width=0.24\textwidth]{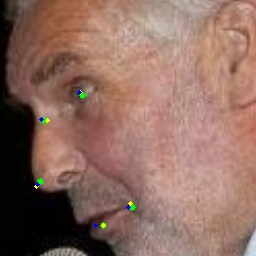}&
\includegraphics[width=0.24\textwidth]{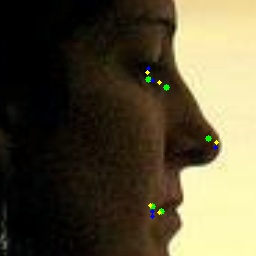}\\
\includegraphics[width=0.24\textwidth]{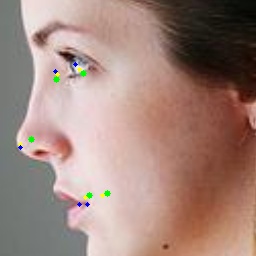}&
\includegraphics[width=0.24\textwidth]{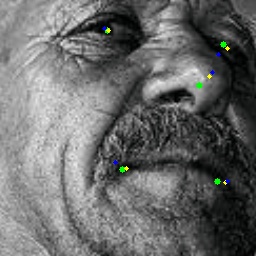}&
\includegraphics[width=0.24\textwidth]{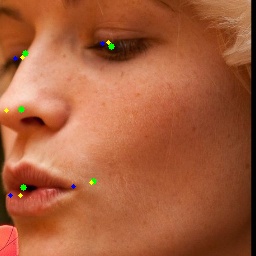}&
\includegraphics[width=0.24\textwidth]{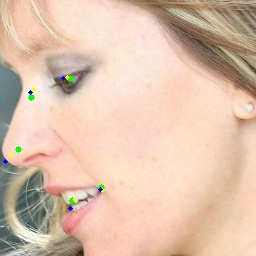}\\
	\end{tabu}
\end{adjustbox}
\caption{\label{tbl:landmarks_pitch_yaw} Five point landmark on images with challenging yaw angles. {\color{blue}Blue}: groundtruth; {\color{green}Green}: baseline; {\color{yellow}Yellow}: with 3D pose augmentation (Best viewed in color).}
\end{center}
\end{figure}

We give qualitative visualization of the results from face landmark localization models trained with and without the proposed 3D augmentation method. Specifically, we show the results of models trained with 3D out-of-plane pose rotation augmentation. \tabref{tbl:landmarks_yaw} and \tabref{tbl:landmarks_pitch_yaw} shows the five landmark locations of left eye, right eye, nose tip, mouth left and mouth right out of the total 68 predicted. We follow the 68 landmark scheme defined in \cite{Sagonas:2016:FIC:2949077.2949200}. We can see that the model trained with 3D-aided pose augmentation is more robust, especially for faces at non-frontal views.

\FloatBarrier
{\small
  \printbibliography
}

%%% Local Variables:
%%% mode: latex
%%% TeX-master: "data_augmentation_for_face_recognition"
%%% End:

\end{document}